**Title**: Common pitfalls and recommendations for using machine learning to detect and prognosticate for COVID-19 using chest radiographs and CT scans

**Authors:** Michael Roberts, Derek Driggs, Matthew Thorpe, Julian Gilbey, Michael Yeung, Stephan Ursprung, Angelica I. Aviles-Rivero, Christian Etmann, Cathal McCague, Lucian Beer, Jonathan R. Weir-McCall, Zhongzhao Teng, Effrossyni Gkrania-Klotsas, James H.F. Rudd*, Evis Sala*, Carola-Bibiane Schönlieb* on behalf of the AIX-COVNET collaboration†

† Members are listed in the appendix.
* Joint senior authors.

**Affiliations:**
**Department of Applied Mathematics and Theoretical Physics** (M Roberts PhD, D Driggs MS, J Gilbey PhD, AI Aviles-Rivero PhD, C Etmann PhD, Prof C-B Schönlieb PhD) **and Department of Radiolog**y (M Yeung BA, S Ursprung MD, C McCague MBBS, L Beer PhD, JR Weir-McCall PhD, Z Teng PhD, Prof Evis Sala PhD) **and Department of Medicine** (JHF Rudd PhD) **and Cancer Research UK Cambridge Centre** (S Ursprung MD, C McCague MBBS, Prof Evis Sala PhD) **at the University of Cambridge, Cambridge, UK**.

**Oncology R&D** (M Roberts PhD) **at AstraZeneca, Cambridge, UK.**

**Department of Mathematics** (M Thorpe PhD) **at the University of Manchester, Manchester, UK.**

**Royal Papworth Hospital** (JR Weir-McCall PhD)**, Cambridge, UK.**

**Department of Infectious Diseases** (Effrossyni Gkrania-Klotsas PhD)**, at Cambridge University Hospitals NHS Trust, Cambridge, UK.**

**Corresponding author:** Michael Roberts at the Department of Applied Mathematics and Theoretical Physics, University of Cambridge, Cambridge, UK (michael.roberts@maths.cam.ac.uk) and Oncology R&D at AstraZeneca, Cambridge, UK (michael.roberts2@astrazeneca.com).

## Abstract

Machine learning methods offer great promise for fast and accurate detection and prognostication of COVID-19 from standard-of-care chest radiographs (CXR) and computed tomography (CT) images. Many articles have been published in 2020 describing new machine learning-based models for both of these tasks, but it is unclear which are of potential clinical utility. In this systematic review, we search EMBASE via OVID, MEDLINE via PubMed, bioRxiv, medRxiv and arXiv for published papers and preprints uploaded from January 1, 2020 to October 3, 2020 which describe new machine learning models for the diagnosis or prognosis of COVID-19 from CXR or CT images. Our search identified 2,212 studies, of which 415 were included after initial screening and, after quality screening, 61 studies were included in this systematic review. Our review finds that none of the models identified are of potential clinical use due to methodological flaws and/or underlying biases. This is a major weakness, given the urgency with which validated COVID-19 models are needed. To address this, we give many recommendations which, if followed, will solve these issues and lead to higher quality model development and well documented manuscripts.

## Introduction

In December 2019, a novel coronavirus was first recognised in Wuhan, China [1]. On January 30, 2020, as infection rates and deaths across China soared and the first death outside China was recorded, the WHO described the then-unnamed disease as a Public Health Emergency of International Concern [2]. The disease was officially named Coronavirus disease 2019 (COVID-19) by February 12, 2020 [3], and was declared a pandemic on March 11, 2020 [4]. Since its first description in late 2019, the COVID-19 infection has spread across the globe, causing massive societal disruption and stretching our ability to deliver effective healthcare. This was caused by a lack of knowledge about the virus's behaviour along with a lack of an effective vaccine and anti-viral therapies.

Although reverse transcription polymerase chain reaction (RT-PCR) is the test of choice for diagnosing COVID-19, imaging can complement its use to achieve greater diagnostic certainty or even be a surrogate in some countries where RT-PCR is not readily available. In some cases, CXR abnormalities are visible in patients who initially had a negative RT-PCR test [5] and several studies have shown that chest CT has a higher sensitivity for COVID-19 than RT-PCR, and could be considered as a primary tool for diagnosis [6–9]. In response to the pandemic, researchers have rushed to develop models using artificial intelligence (AI), in particular machine learning, to support clinicians.

Given recent developments in the application of machine learning models to medical imaging problems [10,11], there is fantastic promise for applying machine learning methods to COVID-19 radiological imaging for improving the accuracy of diagnosis, compared to the gold-standard RT-PCR, whilst also providing valuable insight for prognostication of patient outcomes. These models have the potential to exploit the large amount of multi-modal data collected from patients and could, if successful, transform detection, diagnosis, and triage of patients with suspected COVID-19. Of greatest potential utility is a model which can not only distinguish COVID-19 from non-COVID-19 patients but also discern alternative types of pneumonia such as those of bacterial or other viral aetiologies. With no standardisation, AI algorithms for COVID-19 have been developed with a very broad range of applications, data collection procedures and performance assessment metrics. Perhaps as a result, none are currently ready to be deployed clinically. Reasons for this include: (i) the bias in small data sets; (ii) the variability of large internationally-sourced data sets; (iii) the poor integration of multi-stream data, particularly imaging data; (iv) the difficulty of the task of prognostication, and (v) the necessity for clinicians and data analysts to work side-by-side to ensure the developed AI algorithms are clinically relevant and implementable into routine clinical care. Since the pandemic began in early 2020, researchers have answered the 'call to arms' and numerous machine learning models for diagnosis and prognosis of COVID-19 using radiological imaging have been developed and hundreds of manuscripts have been written. In this paper we reviewed the entire literature of machine learning methods as applied to chest CT and CXR for the diagnosis and prognosis of COVID-19. As this is a rapidly developing field, we reviewed both published and preprint works to ensure maximal coverage of the literature.

While earlier reviews provided a broad analysis of predictive models for COVID-19 diagnosis and prognosis [12–15], this review highlights the unique challenges researchers face when developing classical machine learning and deep learning models using imaging data. This review builds on the approach of Wynants [12]: we assess the risk of bias in the papers considered, going further by incorporating a quality screening stage to ensure only those papers with sufficiently documented methodologies are reviewed in most detail. We also focus our review on the systematic methodological flaws in the current machine learning literature for COVID-19 diagnosis and prognosis models using imaging data. We also give detailed recommendations in five domains: (i) considerations when collating COVID-19 imaging datasets that are to be made public; (ii) methodological considerations for algorithm developers; (iii) specific issues about reproducibility of the results in the literature; (iv) considerations for authors to ensure sufficient documentation of methodologies in manuscripts, and (v) considerations for reviewers performing peer review of manuscripts.

This review has been performed, and informed, by both clinicians and algorithm developers, with our recommendations aimed at ensuring the most clinically relevant questions are addressed appropriately, whilst maintaining standards of practice to help researchers develop useful models and report reliable results even in the midst of a pandemic.

## Results

Study selection: Our initial search highlighted 2,212 papers that satisfied our search criteria; removing duplicates we retained 2,150 papers, and of these, 415 papers had abstracts or titles deemed relevant to the review question, introducing machine learning methods for COVID-19 diagnosis or prognosis using radiological imaging. Full-text screening retained 319 papers, of which, after quality review, 61 were included for discussion in this review (see Figure 1). Of these, 37 were deep learning papers, 22 were traditional machine learning papers and 2 were hybrid papers (using both approaches). The two hybrid papers both failed the CLAIM check but passed RQS.

Quality screening failures:

**Deep learning papers.** There were 254/319 papers which described deep learning based models and 215 of these were excluded from the detailed review (including one hybrid paper). We find that 110 papers (51%) fail at least three of our identified mandatory criteria from the CLAIM checklist (Supp. Mat. A7), with 23% failing two and 26% failing just one. In the rejected papers, the three most common reasons for a paper failing the quality check is due to insufficient documentation of:

(1) how the final model was selected in 61% (132),
(2) the method of pre-processing of the images in 58% (125),
(3) the details of the training approach (e.g. the optimizer, the loss function, the learning rate) in 49% (105).

***Traditional machine learning papers.*** There are 68 papers which describe traditional machine learning methods and 44 of these were excluded from the review, i.e. the RQS is less than 6 or the datasets used are not specified in the paper. There are only two papers which have an RQS $\geq 6$ but which fail to disclose the datasets used in the analysis. Of the remaining papers the two factors which lead to the lowest RQS results are omission of:

(1) feature reduction techniques in 52% of papers (23),
(2) model validation in 61% of papers (27).

Full details can be found in Supp. Mat. A8.

Remaining papers for detailed analysis:

***Deep learning papers.*** There are six non-mandatory CLAIM criteria not satisfied in at least half of the 37 papers:
(1) 29 do not complete any external validation.
(2) 30 do not perform any robustness or sensitivity analysis of their model.
(3) 26 do not report the demographics of their data partitions.
(4) 25 do not report the statistical tests used to assess significance of results or determine confidence intervals.
(5) 23 do not report confidence intervals for the performance.
(6) 22 do not sufficiently report their limitations, biases or issues around generalizability.

The full CLAIM results are in Supp. Mat. A9.

***Traditional machine learning papers.*** Of the 24 papers, including the two hybrid papers, none use longitudinal imaging, perform a prospective study for validation, or standardise image acquisition by using either a phantom study or a public protocol. Only six papers describe performing external validation and only four papers report the calibration statistics (the level of agreement between predicted risks and those observed) and associated statistical significance for the model predictions. The full RQS scores are in Supp. Mat. A9.

Datasets considered:

Public datasets were used extensively in the literature appearing in 32/61 papers (see Supp. Mat. B2 for list of public datasets, three papers use both public and private data). Private data is used in 32/61 papers with 21 using data from mainland China, two using data from France and the remainder using data from Iran, the USA, Belgium, Brazil, Hong Kong and the Netherlands.

Diagnostic models for COVID-19:

**Diagnosis models using CXRs.** Twenty-two papers consider diagnosis of COVID-19 from CXR images [19–39]. Most of these papers use off-the-shelf networks including ResNet-18 or ResNet-50 [19,20,23,29,32,35,40], DenseNet-121 [30,34,35,37,87], VGG-16 or VGG-19 [22,36,38], Inception [24,41] and EfficientNet [33,42], with three considering custom architectures [21,28,39] and three using hand engineered features [25–27]. Most papers classify images into the three classes COVID-19, non-COVID-19 pneumonia and normal [19,22,24,26,28,29,33,35–40,87], while two consider an extra class by dividing non-COVID-19 pneumonia into viral and bacterial pneumonia[20,32]. ResNet and DenseNet architectures reported better performance than the others, with accuracies ranging from 0·88 to 0·99. However, we caution against direct comparison since the papers use different training and testing settings (e.g. different datasets and data partition sizes) and consider a different number of classes.

**Diagnostic models using CT scans and deep learning.** Eighteen papers applied deep learning techniques to CT imaging, all of which were framed as a classification task to distinguish COVID-19 from other lung pathologies such as (viral or bacterial) pneumonia, interstitial lung disease [38,43–50] and/or a non-COVID-19 class [43,44,47,49,51–55]. The full 3D volumes were only considered in seven papers [43,46,50,53,55–57] with the remainder considering isolated 2D slices or even 2D patches [48]. In most 2D models, authors employed transfer learning, with networks pre-trained on ImageNet [58]. Almost all models used lung segmentation as a pre-processing step. One paper [51] used a Generative Adversarial Network (GAN) [59] approach to address the paucity of COVID-19 CT imaging. AUCs are reported ranging from 0·70 to 1·00.

**Diagnostic models using CT scans and traditional machine learning methods**. Eight papers employed traditional machine learning methods for COVID-19 diagnosis using hand-engineered features [43,60–65] or CNN-extracted features [49]. Four papers [49,62,63,65] incorporate clinical features with those obtained from the CT images. All papers using hand-engineered features employed feature reduction, using between 4 and 39 features in their final models. For final classification, five papers used logistic regression [43,61–64], one used a random forest[60], one a multilayer perceptron[49] and one compared many different machine learning classifiers to determine the best [65]. Accuracies ranged from 0·76 to 0·98 [43,49,60–62]. As before, we caution against direct comparison. The traditional machine-learning model in the hybrid paper[43] had a 0·05 lower accuracy than their deep learning model.

Prognostic models for COVID-19 using CT and CXR images: Eighteen papers developed models for the prognosis of patients with COVID-19 [54,66–82], fourteen using CT and four using CXR. These models were developed for predicting severity of outcomes including: death or need for ventilation [80,81], a need for ICU admission[66,75,79–81] , progression to acute respiratory distress syndrome [82], the length of hospital stay [54,83], likelihood of conversion to severe disease [67,68,77] and the extent of lung infection[78]. Most papers used models based on a multivariate Cox proportional hazards model [54,80,81], logistic regression [68,75–77,82,83], linear regression [77,78], random forest [79,83] or compare a huge variety of machine learning models such as tree-based methods, support vector machines, neural networks and nearest neighbour clustering [66,67].

Predictors from radiological data were extracted using either handcrafted radiomic features [66,67,71–73,77,79–83] or deep learning [54,69,73–75,78]. Clinical data included basic observations, serology and comorbidities. Only seven models integrated both radiological and clinical data [65,66,72,75,79–81].

Risks of bias: Following the PROBAST guidance, the risk of bias was assessed for all 61 papers in four domains: participants, predictors, outcomes and analysis; the results are shown in Table 1. We find that 54/61 papers had a high risk of bias in at least one domain with the others unclear in at least one domain.

***Participants.*** Almost all papers had a high (45/61) or unclear (10/61) risk of bias for their participants, with only six assessed as having a low risk of bias. This is primarily due to the following issues: (i) for public datasets it is not possible to know whether patients are truly COVID-19 positive, or if they have underlying selection biases, as anybody can contribute images [19,27,29,31–35,37,38,40,44,47,51,52,78,87]; (ii) the paper uses only a subset of original datasets, applying some exclusion criteria, without enough details to be reproducible [19,46,47,51,52,54,64,73,74,77,78], and/or (iii) there are significant differences in demographics between the COVID-19 cohort and the control groups, with e.g. paediatric patients as controls [20,27,31,32,34,35,38,40,48,49,62,84,87].

***Predictors.*** For models where the features have been extracted using deep learning models, the predictors are unknown and abstract imaging features. Therefore, for these papers (38/61), we cannot judge biases in the predictors. For 19 papers, the risk of bias is recorded as low due to the use of pre-defined hand-engineered features. For the remaining 4 papers, a high risk of bias is recorded due to the predictors being assessed with knowledge of the associated outcome.

***Outcomes*.** The risk of bias in the outcome variable was found to be low for the majority (24/61) of the papers, unclear for 26/61 and high for 11/61. To evaluate the bias in the outcome, we took different approaches for papers using private datasets and public datasets (three papers use a mixture).

For the 32 papers that use public datasets, the outcome was assigned by the originators of the dataset and not by the papers' authors. Papers using a public dataset generally have an unclear risk of bias (27/32) as they have used the outcome directly sourced from the dataset originator.

For the 32 papers that use private datasets, the COVID-19 diagnosis is due to either positive RT-PCR or antibody tests for 23/32 and have a low risk of bias. The other papers have a high (7/32) or unclear (2/32) risk of bias due to inconsistent diagnosis of COVID-19 [18,43], unclear definition of a control group [66,68], ground truths being assigned using the images themselves [29,57,63,74], using an unestablished reference to define outcome [83] or by combining public and private datasets [44,69,85].

***Analysis*.** Only ten papers have a low risk of bias for their analysis. The high risk of bias in most papers is principally due to a small sample size of COVID-19 patients (leading to highly imbalanced datasets), use of only a single internal holdout set for validating their algorithm (rather than cross-validation or bootstrapping) and a lack of appropriate evaluation of the performance metrics (e.g. no discussion of calibration/discrimination)[21–23,25,26,47,51,55,67,82]. One paper with a high risk of bias[35] claims external validation on dataset [16], not realising that this already includes both datasets [17] and [86] that were used to train the algorithm.

<u>Data analysis:</u> There are two approaches for validating the performance of an algorithm, namely internal and external validation. For internal validation, the test data is from the same source as the development data and for external validation they are from different sources. Including both internal and external validation allows more insight to generalisability of the algorithm. We find 47/61 papers consider internal validation only with 13/61 using external validation [25,35,44,45,54,57,66,69,70,72,75,80,81]. Twelve used truly external test datasets and one tested on the same data the algorithm was trained on[35].

***Model evaluation.*** In Table 2 we give the performance metrics quoted in each paper. Ten papers use cross-validation to evaluate model performance [24,38,39,50,52,60,68,77,79,83], one uses both cross-validation and an external test set [44], one quotes correlation metrics [78] and one has an unclear validation method [20]. The other papers all have an internal holdout or external test set with sensitivity and specificity derived from the test data using an unquoted operating point (with exception of [19] that quotes operating point 0·5). It would be expected that an operating point be chosen based on the algorithm performance for the validation data used to tune and select the final algorithm. However, the ROC curves and AUC values are given for the internal holdout or external test data independent of the validation data.

***Partition analysis.*** In Figure 2, we show the quantity of data (split by class) used in the training cohort of 32 diagnosis models. We exclude many studies [21,23,61,74,87,25,26,28,32,35,38,46,48] because it was unclear how many images were used. If a paper only stated the number of patients (and not the number of images), we assumed that there was only one image per patient. We see that 20/32 papers have a reasonable balance between classes (with exceptions being [20,27,29,33,34,36,39,40,43,54,64,65]. However, the majority of datasets are quite small, with 19/32 papers using fewer than 2,000 datapoints for development (with exceptions [20,29,30,33,34,36,39,44,51,56,57,60,84]. Only seven papers used both a dataset with more than 2,000 datapoints that was balanced for COVID-19 positive and the other classes [30,44,51,56,57,60,84].

Figure 3 shows the number of images of each class used in the holdout/test cohorts. We find 6/32 papers had an imbalanced testing dataset [20,27,36,39,40,64] Only 6/32 papers tested on more than 1,000 images [20,30,39,44,57,84]. Only 4/32, had both a large and balanced testing dataset [30,44,57,84].

<u>Public availability of the algorithms and models:</u> Only 13/61 papers [24,26,30,33,37,39,49,54,69,70,78,83,84] publish the code for reproducing their results (seven including their pre-trained parameters) and one states that it is available on request [83].

## Discussion

Our systematic review highlights the extensive efforts of the international community to tackle the COVID-19 pandemic using machine learning. These early studies show promise for diagnosis and prognostication of pneumonia secondary to COVID-19. However, we have also found that current reports suffer from a high prevalence of deficiencies in methodology and reporting, with none of the reviewed literature reaching the threshold of robustness and reproducibility essential to support utilisation in clinical practice. Many studies are hampered by issues with poor quality data, poor application of machine learning methodology, poor reproducibility, and biases in study design. The current paper complements the work of Wynants et al. who have published a living systematic review [12] on publications and preprints of studies describing multivariable models for screening of COVID-19 infections in the general population, differential-diagnosis of COVID-19 infection in symptomatic patients, and prognostication in patients with confirmed COVID-19 infection. While Wynants et al. reviewed multivariable models with any type of clinical input data, the present review focuses specifically on machine learning based diagnostic and prognostic models using medical imaging. Furthermore, this systematic review employed specialised quality metrics for the assessment of radiomics and deep learning-based diagnostic models in radiology. This is also in contrast to previous studies that have assessed AI algorithms in COVID-19 [13,14]. Limitations of the current literature most frequently reflect either a limitation of the dataset used in the model or methodological mistakes repeated in many studies that likely lead to overly optimistic performance evaluations.

Datasets: Many papers gave little attention to establishing the original source of the images (Supp. Mat. B2). When considering papers that use public data, readers should be aware of the following:

***Duplication and quality issues.*** There is no restriction for a contributor to upload COVID-19 images to many of the public repositories [17,88–91]. There is high likelihood of duplication of images across these sources and no assurance that the cases included in these datasets are confirmed COVID-19 cases (authors take a great leap to assume this is true) so great care must be taken when combining datasets from different public repositories. Also, most of the images have been pre-processed and compressed into non-DICOM formats leading to a loss in quality and a lack of consistency/comparability.

- ***Source issues.*** Many papers (16/61) use the pneumonia dataset of Kermany et al. [86] as a control group. They commonly fail to mention that this consists of paediatric patients aged between one and five. Developing a model using adult COVID-19 patients and very young pneumonia patients is likely to overperform as it is merely detecting children vs. adults. This dataset is also erroneously referred to as the Mooney dataset in many papers (being the Kermany dataset deployed on Kaggle [92]). It is also important to consider the sources of each image class, for example if images for different diagnoses are from different sources. It is demonstrated by Maguolo et al. [93] that by excluding the lung region entirely, the authors could identify the source of the Cohen [17], Kermany [86] an AUC between 0.9210 to 0.9997 and 'diagnose' COVID-19 with an AUC=0·68.

- ***Frankenstein datasets.*** The issues of duplication and source become compounded when public 'Frankenstein' datasets are used, that is, datasets assembled from other datasets and redistributed under a new name. For instance, one dataset [92] combines several other datasets [17,89,94] without realising that one of the component datasets [94] already contains another component [89]. This repackaging of datasets, although pragmatic, inevitably leads to problems with algorithms being trained and tested on identical or overlapping datasets whilst believing them to be from distinct sources.

- ***Implicit biases in the source data.*** Images uploaded to a public repository and those extracted from publications [94] are likely to have implicit biases due to the contribution source. For example, it is likely that more interesting, unusual or severe cases of COVID-19 appear in publications.

Methodology: All proposed models suffer from a high or unclear risk of bias in at least one domain. There are several methodological issues driven by the urgency in responding to the COVID-19 crisis and subtler sources of bias due to poor application of machine learning.

The urgency of the pandemic led to many studies using datasets that contain obvious biases or are not representative of the target population, e.g. paediatric patients. Before evaluating a model, it is crucial that authors report the demographic statistics for their datasets, including age and sex distributions. Diagnostic studies commonly compare their models' performance to that of RT-PCR. However, as the ground-truth labels are often determined by RT-PCR, there is no way to measure whether a model outperforms RT-PCR from accuracy, sensitivity, or specificity metrics alone. Ideally, models should aim to match clinicians using all available clinical and radiomic data, or to aid them in decision making.

Many papers utilise transfer learning in developing their model, which assumes an inherent benefit to performance. However, it is unclear whether transfer learning offers significant performance benefit due to the over-parametrisation of the models [44,61]. Many publications used the same resolutions such as 224-by-224 or 256-by-256 for training, which are often used for ImageNet classification, indicating that the pre-trained model dictated the image rescaling used rather than clinical judgement.

Recommendations: Based on the systematic issues we encountered in the literature, we offer recommendations in five distinct areas: (i) the data used for model development and common pitfalls; (ii) the evaluation of trained models; (iii) reproducibility; (iv) documentation in manuscripts, and (v) the peer review process. Our recommendations in areas (iii) and (iv) are largely informed by the 258 papers that did not pass our initial quality check, while areas (i), (ii) and (v) follow from our analysis of the 61 papers receiving our full review.

**Recommendations for data.** Firstly, we advise caution over the use of public repositories, which can lead to significant risks of bias due to source issues and Frankenstein datasets as discussed above.. Furthermore, authors should aim to match demographics across cohorts, an often neglected but significant potential source of bias; this can be impossible with public datasets that do not include demographic information, and including paediatric images [86] in the COVID-19 context introduces a strong bias.

Using a public dataset alone without additional new data can lead to community-wide overfitting on this dataset. Even if each individual study observes sufficient precautions to avoid overfitting, the fact that the community is focused on outperforming benchmarks on a single public dataset encourages overfitting. Many public datasets containing images taken from preprints receive these images in low-resolution or compressed formats (e.g. JPEG and PNG), rather than their original DICOM format. This loss of resolution is a serious concern for traditional machine learning models if the loss of resolution is not uniform across classes, and the lack of DICOM metadata does not allow exploration of model dependence on image acquisition parameters (e.g. scanner manufacturer, slice thickness, etc.).

Regarding CXRs, researchers should be aware that algorithms might associate more severe disease not with CXR imaging features, but the view that has been used to acquire that CXR. For example, in sick, immobile patients, an anteroposterior CXR view is used for practicality rather than the standard posteroanterior CXR projection. Also, overrepresentation of severe disease is not only bad from the machine learning perspective, but also in terms of clinical utility, since the most useful algorithms are those that can diagnose disease at an early stage [95]. The timing between imaging and RT-PCR tests was also largely undocumented, which has implications for the validity of the ground truth used. It is also important to recognise that a negative RT-PCR test does not necessarily mean that a patient does not have COVID-19. We encourage authors to evaluate their algorithms on datasets from the pre-COVID-19 era, such as performed by [96], to validate any claims that the algorithm is isolating COVID-19-specific imaging features. It is common for non-COVID-19 diagnoses (for example, non-COVID-19 pneumonia) to be determined from imaging alone. However, in many cases these images are the only predictors of the developed model, and using predictors to inform outcomes leads to optimistic performance.

**Recommendations for evaluation.** We emphasise the importance of using a well-curated external validation dataset of appropriate size in order to assess generalizability to other cohorts. Any useful model for diagnosis or prognostication must be robust enough to give reliable results for any sample from the target population rather than just on the sampled population. Calibration statistics should be calculated for the developed models to inform predictive error and decision curve analysis [97] performed for assessing clinical utility. It is important for authors to state how they ensured that images from the same patient were not included in the different dataset partitions, such as describing patient-level splits. This is an issue for approaches that consider 2D and 3D images as a single sample and also for those which process 3D volumes as independent 2D samples. It is also important when using datasets containing multiple images from each patient. When reporting results, it is important to include confidence intervals to reflect the uncertainty in the estimate, especially when training models on the small sample sizes commonly seen with COVID-19 data. Moreover, we stress the importance of not only reporting results, but also demonstrating model interpretability with methods such as saliency maps, which is a necessary consideration for adoption into clinical practice. We remind authors that it is inappropriate to compare model performance to RT-PCR or any other ground truths. Instead, authors should aim for models to either improve the performance and efficiency of clinicians, or, even better, to aid clinicians by providing interpretable predictions. Examples of interpretability techniques include: (i) informing the clinician of which features in the data most influenced the prediction of the model, (ii) linking the prognostic features to the underlying biology and (iii) overlaying an activation/saliency map on the image to indicate the region of the image which influenced the model's prediction and (iv) identifying patients which had a similar clinical pathway.

Most papers derive their performance metrics from the test data alone with an unstated operating point to calculate sensitivity and specificity. Clinical judgment should be used to identify the desired sensitivity or specificity of the model and the operating point should be derived from the development data. The differences in the sensitivity and specificity of the model should be recorded separately for the validation and test data. Using an operating point of 0·5 and only reporting the test sensitivity and specificity fails to convey the reliability of the threshold. This is a key aspect of generalisability. Omitting it would see an FDA 510K submission rejected.

**Recommendations for replicability.** A possible ambiguity arises due to updating of publicly available datasets or code. Therefore, we recommend that a cached version of the public dataset be saved, or the date/version quoted, and specific versions of data or code be appropriately referenced. (Git commit ids or tags can be helpful for this purpose to reference a specific version on GitHub, for example.) We acknowledge that although perfect replication is potentially not possible, details such as the seeds used for randomness and the actual partitions of the dataset for training, validation and testing would form very useful supplementary materials.

**Recommendations for authors.** For authors, we recommend assessing their paper against appropriate established frameworks, such as RQS, CLAIM, TRIPOD, PROBAST and QUADAS [98–102]. By far the most common point leading to exclusion was failure to state the data pre-processing techniques in sufficient detail. As a minimum, we expected papers to state any image resizing, cropping and normalisation used prior to model input, and with this small addition many more papers would have passed through the quality review stage. Other commonly missed points include details of the training (such as number of epochs and stopping criteria), robustness or sensitivity analysis, and the demographic or clinical characteristics of patients in each partition.

**Recommendations for reviewers.** For reviewers, we also recommend the use of the checklists [98–102] in order to better identify common weaknesses in reporting the methodology. The most common issues in the papers we reviewed was the use of biased datasets and/or methodologies. For non-public datasets, it may be difficult for reviewers to assess possible biases if an insufficiently detailed description is given by the authors. We strongly encourage reviewers to ask for clarification from the authors if there is any doubt about bias in the model being considered. Finally, we suggest using reviewers from a combination of both medical and machine learning backgrounds, as they can judge the clinical and technical aspects in different ways.

<u>Challenges and opportunities:</u> Models developed for diagnosis and prognostication from radiological imaging data are limited by the quality of their training data. While many public datasets exist for researchers to train deep learning models for these purposes, we have determined that these datasets are not large enough, or of suitable quality, to train reliable models, and all studies using publicly available datasets exhibit a high or unclear risk of bias. However, the size and quality of these datasets can be continuously improved if researchers world-wide submit their data for public review. Because of the uncertain quality of many COVID-19 datasets, it is likely more beneficial to the research community to establish a database which has a systematic review of submitted data than it is to immediately release data of questionable quality as a public database.

The intricate link of any AI algorithm for detection, diagnosis or prognosis of COVID-19 infections to a clear clinical need is essential for successful translation. As such, complementary computational and clinical expertise, in conjunction with high quality healthcare data, are required for the development of AI algorithms. Meaningful evaluation of an algorithm's performance is most likely to occur in a prospective clinical setting. Like the need for collaborative development of AI algorithms, the complementary perspectives of experts in machine learning and academic medicine were critical in conducting this systematic review.

<u>Limitations:</u> Due to the fast development of diagnostic and prognostic AI algorithms for COVID-19, at the time of finalising our analyses, several new preprints have been released; these are not included in this study.

Our study has limitations in terms of methodologic quality and exclusion. Several high-quality papers published in high-impact journals - including Radiology, Cell and IEEE Transactions on Medical Imaging - were excluded due to the lack of documentation on the proposed algorithmic approaches. As the AI algorithms are the core for the diagnosis and prognosis of COVID-19, we only included works that are reproducible. Furthermore, we acknowledge that the CLAIM requirements are harder to fulfil than the RQS ones, and the paper quality check is therefore not be fully comparable between the two. We underline that several excluded papers were preprint versions and may possibly pass the systematic evaluation in a future revision.

In our PROBAST assessment, for the "were there a reasonable number of participants?" question of the analysis domain, we required a model to be trained on at least 20 events-per-variable for the size of the dataset to score a low risk of bias [101]. However, events-per-variable may not be a useful metric to determine if a deep learning model will overfit. Despite their gross over-parameterisation, deep learning models generalise well in a variety of tasks, and it is difficult to determine *a priori* whether a model will overfit given the number of training examples [103]. A model which was trained using less than 500 COVID-19 positive images was deemed to have a high risk of bias in answer to this and more than 2000 COVID-19 positive images qualified as low risk. However, in determining the overall risk of bias for the analysis domain we factor in nine PROBAST questions, so it is possible for a paper using less than 500 images to achieve at best an unclear overall risk of bias for its analysis. Similarly, it is possible for papers which have over 2000 images to have an overall high risk of bias for their analysis if it does not account for other sources of bias.

## Conclusions

This is the first systematic review to specifically consider the current machine learning literature using CT and CXR imaging for COVID-19 diagnosis and prognosis which emphasises the quality of the methodologies applied and the reproducibility of the methods. We found that no papers in the literature currently have all of: (i) a sufficiently documented manuscript describing a reproducible method; (ii) a method which follows best practice for developing a machine learning model, and (iii) sufficient external validation to justify the wider applicability of the method. We give detailed significant recommendations for data curators, machine learning researchers, manuscript authors and reviewers to ensure the best quality methods are developed which are reproducible and free from biases in either the underlying data or the model development.

Despite the huge efforts of researchers to develop machine learning models for COVID-19 diagnosis and prognosis, we find methodological flaws and significant biases throughout the literature, leading to highly optimistic reported performance. In their current reported form, none of the machine learning models included in this review are likely candidates for clinical translation for the diagnosis/prognosis of COVID-19. Higher quality datasets, manuscripts with sufficient documentation to be reproducible and external validation are required to increase the likelihood of models being taken forward and integrated into future clinical trials to establish independent technical and clinical validation as well as cost-effectiveness.

## Methods

The methods for performing this systematic review are registered with PROSPERO [CRD42020188887] and were agreed by all authors before the start of the review process, to avoid bias.

Search strategy and selection criteria: We have followed the PRISMA checklist [104] and include this in Supp. Mat. C. We performed our search to identify published and unpublished works using the arXiv and the "Living Evidence on COVID-19" database [105], a collation of all COVID-19 related papers from EMBASE via OVID, MEDLINE via PubMed, bioRxiv and medRxiv. The databases were searched from Jan 1, 2020 through to October 3, 2020. The full search strategy is detailed in the appendix. The initial cut-off is chosen to specifically include all early COVID-19 research, given that the World Health Organisation was only informed of the "pneumonia of unknown cause" on Dec 31, 2019 [106]. An initial search was performed on May 28, 2020, with updated searches performed on June 24, 2020, August 14, 2020, August 15, 2020 and 3 October 2020 to identify any relevant new papers published in the intervening period. Since many of the papers identified are preprints, some of them were updated or published between these dates; in such cases, we used the preprint as it was at the later search date or the published version. Some papers were identified as duplicates ourselves or by Covidence [107]; in these instances we ensured that the latest version of the paper was reviewed. We used a three-stage process to determine which papers would be included in this review. During the course of the review, one author (A. A.-R.) submitted a paper [108] which in scope for this review, however we excluded it due to the potential for conflict of interest.

***Title and abstract screening***. In the first stage, a team of ten reviewers assessed papers for eligibility, screening the titles and abstracts to ensure relevance. Each paper was assessed by two reviewers independently and conflicts were resolved by consensus of the ten reviewers (see Supp. Mat. A4).

***Full-text screening***. In the second stage, the full text of each paper was screened by two reviewers independently to ensure that the paper was eligible for inclusion with conflicts resolved by consensus of the ten reviewers.

***Quality review.*** In the third stage, we considered the quality of the documentation of methodologies in the papers. Note that exclusion at this stage is not a judgement on the quality or impact of a paper or algorithm, merely that the methodology is not documented with enough detail to allow the results to be reliably reproduced.

At this point we separated machine learning methods into deep learning methods and non-deep learning methods (we refer to these as traditional machine learning methods). The traditional machine learning papers were scored using the Radiomic Quality Score (RQS) of Lambin et al. [98], while the deep learning papers were assessed against the Checklist for Artificial Intelligence in Medical Imaging (CLAIM) of Mongan et al. [99]. The ten reviewers were assigned to five teams of two: four of the ten reviewers have a clinical background and were paired with non-clinicians in four of the five teams to ensure a breadth of experience when reviewing these papers. Within each team, the two reviewers independently assessed each paper against the appropriate quality measure. Where papers contained both deep learning and traditional machine learning methodologies, these were assessed using both CLAIM and RQS. Conflicts were resolved by a third reviewer.

To restrict consideration to only those papers with the highest quality documentation of methodology, we excluded papers that did not fulfil particular CLAIM or RQS requirements. For the deep learning papers evaluated using the CLAIM checklist we selected eight checkpoint items deemed mandatory to allow reproduction of the paper's method and results. For the traditional machine learning papers, evaluated using the RQS, we used a threshold of 6 points out of 36 for inclusion in the review along with some basic restrictions, such as detail of the data source and how subsets were selected. The rationale for these CLAIM and RQS restrictions is given in Supp. Mat. A7. If a paper was assessed using both CLAIM and RQS then it only needed to pass one of the quality checks to be included.

In a number of cases, various details of pre-processing, model configuration or training setup were not discussed in the paper, even though they could be inferred from a referenced online code repository (typically GitHub). In these cases, we have assessed the papers purely on the content in the paper, as it is important to be able to reproduce the method and results independently of the authors' code.

Risk of bias in individual studies: We use the Prediction model Risk Of Bias Assessment Tool (PROBAST) of Wolff et al. [101] to assess bias in the datasets, predictors and model analysis in each paper. The papers that passed the quality assessment stage were split amongst three teams of two reviewers to complete the PROBAST review. Within each team, the two reviewers independently scored the risk of bias for each paper and then resolved by conflicts any remaining conflicts were resolved by a third reviewer.

Data analysis: The papers were allocated amongst five teams of two reviewers. These reviewers independently extracted the following information: (i) whether the paper described a diagnosis or prognosis model; (ii) the data used to construct the model; (iii) whether there were predictive features used for the model construction; (iv) the sample sizes used for the development and holdout cohorts (along with the number of COVID-19 positive cases); (v) the type of validation performed; (vi) the best performance quoted in the paper for the validation cohort (whether internal, external or both), and (vii) whether the code for training the model and the trained model were publicly available. Any conflicts were initially resolved by team discussions and remaining conflicts were resolved by a third reviewer.

Role of the funding source: The funders of the study had no role in the study design, data collection, data analysis, data interpretation, or writing of the manuscript. All authors had full access to all the data in the study and had final responsibility for the decision to submit for publication.

**Funding**

There is no direct funding source for this study, however the following authors do report indirect funding as follows:

AstraZeneca (M.R.), European Research Council under the European Union's Horizon 2020 research and innovation programme grant agreement No 777826 (M.T.), Wellcome Trust (C.E, C.M., J.R.), The Mark Foundation for Cancer

Research (L.B., E.S.), Cancer Research UK Cambridge Centre [C9685/A25177] (L.B., E.S., C-B.S.), British Heart Foundation (Z.T., J.R.), the NIHR Cambridge Biomedical Research Centre (Z.T., J.R.), HEFCE (J.R.), Lyzeum Ltd. (J.G.), The Gates Cambridge Trust (D.D.), the Cambridge International Trust (S.U.), the EPSRC Centre Nr. EP/N014588/1 (A.A-R., J.R., C-B.S.) and the Cantab Capital Institute for the Mathematics of Information (A.A-R, C-B.S.)

Additionally, C-B.S. acknowledges support from the Leverhulme Trust project on 'Breaking the non-convexity barrier', the Philip Leverhulme Prize, the EPSRC grants EP/S026045/1 and EP/T003553/1, the Wellcome Innovator Award RG98755, European Union Horizon 2020 research and innovation programmes under the Marie Skłodowska-Curie grant agreement No. 777826 NoMADS and No. 691070 CHiPS and the Alan Turing Institute.

All authors had full access to all the data in the study and the corresponding author had final responsibility for the decision to submit for publication.

**Additional Information:**

**Supplementary information** for this paper is available at https://doi.org/10.17863/CAM.61926.

**Correspondence** should be addressed to M.R.

**Acknowledgments**

M.R. would like to acknowledge the support of his supervisors Mishal Patel and Andrew Reynolds at AstraZeneca. The AIX-COV-NET collaboration would also like to acknowledge the support of Intel in supporting the collaboration.

**Author contributions**

M.R., D.D., M.T., J.G., M.Y., A.A-R., C.E., C.M., S.U., L.B. contributed to the literature search, screening, quality assessment. M.R., D.D., M.T., J.G., M.Y., A.A-R., C.E., C.M., S.U. contributed to the data extraction, analysis and interpretation along with writing the original draft manuscript. M.R., D.D., M.T., J.G., M.Y., A.A-R., C.E., C.M., S.U., L.B., J. W-M., Z.T., E.G.-K, J.R., E.S. and C-B.S. contributed critical revisions of the manuscript, and all authors approved the final manuscript.

**Declaration of interests**

M.R., D.D., M.T., J.G., M.Y., A.A-R., C.E., C.M., S.U., E.G.-K., L.B., J.W-M., Z.T., J.R., E.S. and C-B.S. have no conflicts of interest to declare.

**References**

1. Wang, D. *et al.* Clinical Characteristics of 138 Hospitalized Patients with 2019 Novel Coronavirus-Infected Pneumonia in Wuhan, China. *JAMA - J. Am. Med. Assoc.* **323**, 1061–1069 (2020).

2. Zheng, Y. Y., Ma, Y. T., Zhang, J. Y. & Xie, X. COVID-19 and the cardiovascular system. *Nature Reviews Cardiology* **17**, 259–260 (2020).

3. WHO. WHO Director-General's remarks at the media briefing on 2019-nCoV on 11 February 2020. *World Health Organization, Geneva* (2020).

4. WHO. WHO Director-General's opening remarks at the media briefing on COVID-19 - 11 March 2020. *World Health Organization, Geneva* (2020).

5. Wong, H. Y. F. *et al.* Frequency and Distribution of Chest Radiographic Findings in COVID-19 Positive Patients. *Radiology* **296**, 201160 (2019).

6. Long, C. *et al.* Diagnosis of the Coronavirus disease (COVID-19): rRT-PCR or CT? *Eur. J. Radiol.* **126**, (2020).

7. Fang, Y. *et al.* Sensitivity of Chest CT for COVID-19: Comparison to RT-PCR. *Radiology* **296**, 200432 (2020).

8. Ai, T. *et al.* Correlation of Chest CT and RT-PCR Testing in Coronavirus Disease 2019 (COVID-19) in China: A Report of 1014 Cases. *Radiology* 200642 (2020). doi:10.1148/radiol.2020200642

9. Sperrin, M., Grant, S. W. & Peek, N. Prediction models for diagnosis and prognosis in Covid-19. *The BMJ* **369**, (2020).

10. Irvin, J. *et al.* CheXpert: A Large Chest Radiograph Dataset with Uncertainty Labels and Expert Comparison. *Proc. AAAI Conf. Artif. Intell.* **33**, 590–597 (2019).

11. Huang, P. *et al.* Prediction of lung cancer risk at follow-up screening with low-dose CT: a training and validation study of a deep learning method. *Lancet Digit. Heal.* **1**, e353–e362 (2019).

12. Wynants, L. *et al.* Prediction models for diagnosis and prognosis of covid-19: Systematic review and critical appraisal. *BMJ* **369**, (2020).

13. Hamzeh, A. *et al.* Artificial intelligence techniques for Containment COVID-19 Pandemic: A Systematic Review. *Res. Sq.* (2020). doi:https://doi.org/10.21203/rs.3.rs-30432/v1

14. Albahri, O. S. *et al.* Systematic review of artificial intelligence techniques in the detection and classification of COVID-19 medical images in terms of evaluation and benchmarking: Taxonomy analysis, challenges, future solutions and methodological aspects. *Journal of Infection and Public Health* (2020). doi:10.1016/j.jiph.2020.06.028

15. Feng, S. *et al.* Review of artificial intelligence techniques in imaging data acquisition, segmentation and diagnosis for covid-19. *IEEE Rev. Biomed. Eng.* (2020).

16. COVID-19 Radiography Database | Kaggle. Available at: https://www.kaggle.com/tawsifurrahman/covid19-radiography-database. (Accessed: 29th July 2020)

17. Cohen, J. P., Morrison, P. & Dao, L. COVID-19 Image Data Collection. *arXiv* (2020). Available at: http://arxiv.org/abs/2003.11597. (Accessed: 29th July 2020)

18. Yang, X. *et al.* COVID-CT-Dataset: A CT Scan Dataset about COVID-19. *arXiv* (2020). Available at: http://arxiv.org/abs/2003.13865. (Accessed: 30th July 2020)

19. Tartaglione, E., Barbano, C. A., Berzovini, C., Calandri, M. & Grangetto, M. Unveiling COVID-19 from chest x-ray with deep learning: A hurdles race with small data. *Int. J. Environ. Res. Public Health* **17**, 1–17 (2020).

20. Ghoshal, B. & Tucker, A. Estimating Uncertainty and Interpretability in Deep Learning for Coronavirus (COVID-19) Detection - v2. *arXiv* (2020). Available at: http://arxiv.org/abs/2003.10769. (Accessed: 29th July 2020)

21. Malhotra, A. *et al.* Multi-Task Driven Explainable Diagnosis of COVID-19 using Chest X-ray Images. (2020).

22. Rahaman, M. M. *et al.* Identification of COVID-19 samples from chest X-Ray images using deep learning: A comparison of transfer learning approaches. *J. Xray. Sci. Technol.* **28**, 821–839 (2020).

23. Amer, R., Frid-Adar, M., Gozes, O., Nassar, J. & Greenspan, H. COVID-19 in CXR: from Detection and Severity Scoring to Patient Disease Monitoring. (2020).

24. Tsiknakis, N. *et al.* Interpretable artificial intelligence framework for COVID-19 screening on chest X-rays. *Exp. Ther. Med.* **20**, 727–735 (2020).

25. Elaziz, M. A. *et al.* New machine learning method for imagebased diagnosis of COVID-19. *PLoS One* **15**, (2020).

26. Gil, D., Díaz-Chito, K., Sánchez, C. & Hernández-Sabaté, A. Early Screening of SARS-CoV-2 by Intelligent Analysis of X-Ray Images. (2020).

27. Tamal, M. *et al.* An Integrated Framework with Machine Learning and Radiomics for Accurate and Rapid Early Diagnosis of COVID-19 from Chest X-ray. *medRxiv* (2020). doi:10.1101/2020.10.01.20205146

28. Bararia, A., Ghosh, A., Bose, C., Bhar, D. & Author, C. Network for subclinical prognostication of COVID

19 Patients from data of thoracic roentgenogram: A feasible alternative screening technology. *medRxiv* (2020). doi:10.1101/2020.09.07.20189852

29. Wang, Z. *et al.* Automatically discriminating and localizing COVID-19 from community-acquired pneumonia on chest X-rays. *Pattern Recognit.* **110**, 107613 (2021).

30. Zhang, R. *et al.* Diagnosis of COVID-19 Pneumonia Using Chest Radiography: Value of Artificial Intelligence. *Radiology* 202944 (2020). doi:10.1148/radiol.2020202944

31. Ezzat, D., Hassanien, A. ell & Ella, H. A. GSA-DenseNet121-COVID-19: a Hybrid Deep Learning Architecture for the Diagnosis of COVID-19 Disease based on Gravitational Search Optimization Algorithm. *arXiv* (2020). Available at: http://arxiv.org/abs/2004.05084. (Accessed: 29th July 2020)

32. Farooq, M. & Hafeez, A. COVID-ResNet: A Deep Learning Framework for Screening of COVID19 from Radiographs. *arXiv* (2020).

33. Luz, E. *et al.* Towards an Effective and Efficient Deep Learning Model for COVID-19 Patterns Detection in X-ray Images - v4. *arXiv* (2020). Available at: http://arxiv.org/abs/2004.05717. (Accessed: 29th July 2020)

34. Bassi, P. R. A. S. & Attux, R. A Deep Convolutional Neural Network for COVID-19 Detection Using Chest X-Rays - v2. *arXiv* (2020). Available at: http://arxiv.org/abs/2005.01578. (Accessed: 29th July 2020)

35. Gueguim Kana, E. B., Zebaze Kana, M. G., Donfack Kana, A. F. & Azanfack Kenfack, R. H. A web-based Diagnostic Tool for COVID-19 Using Machine Learning on Chest Radiographs (CXR). *medRxiv* (2020). doi:10.1101/2020.04.21.20063263

36. Heidari, M. *et al.* Improving the performance of CNN to predict the likelihood of COVID-19 using chest X-ray images with preprocessing algorithms. *Int. J. Med. Inform.* **144**, (2020).

37. Li, X., Li, C. & Zhu, D. COVID-MobileXpert: On-Device COVID-19 Screening using Snapshots of Chest X-Ray - v3. *arXiv* (2020). Available at: http://arxiv.org/abs/2004.03042. (Accessed: 30th July 2020)

38. Zokaeinikoo, M., Mitra, P., Kumara, S. & Kazemian, P. AIDCOV: An Interpretable Artificial Intelligence Model for Detection of COVID-19 from Chest Radiography Images - v3. *medRxiv* (2020). doi:10.1101/2020.05.24.20111922

39. Sayyed, A. Q. M. S., Saha, D. & Hossain, A. R. CovMUNET: A Multiple Loss Approach towards Detection of COVID-19 from Chest X-ray. (2020).

40. Zhang, R. *et al.* COVID19XrayNet: A Two-Step Transfer Learning Model for the COVID-19 Detecting Problem Based on a Limited Number of Chest X-Ray Images. *Interdiscip. Sci. Comput. Life Sci.* **12**, 555–565 (2020).

41. Szegedy, C., Vanhoucke, V., Ioffe, S., Shlens, J. & Wojna, Z. Rethinking the Inception Architecture for Computer Vision. in *Proceedings of the IEEE Computer Society Conference on Computer Vision and Pattern Recognition* **2016**-**Decem**, 2818–2826 (IEEE Computer Society, 2016).

42. Tan, M. & Le, Q. EfficientNet: Rethinking Model Scaling for Convolutional Neural Networks. in *International Conference of Machine Learning ICML* (2019).

43. Georgescu, B. *et al.* Machine Learning Automatically Detects COVID-19 using Chest CTs in a Large Multicenter Cohort - v2. *arXiv* (2020). Available at: http://arxiv.org/abs/2006.04998. (Accessed: 29th July 2020)

44. Ko, H. *et al.* COVID-19 Pneumonia Diagnosis Using a Simple 2D Deep Learning Framework With a Single Chest CT Image: Model Development and Validation. *J. Med. Internet Res.* **22**, e19569 (2020).

45. Wang, S. *et al.* A deep learning algorithm using CT images to screen for corona virus disease (COVID-19) - v5. *medRxiv* 1–19 (2020). doi:10.1101/2020.02.14.20023028

46. Pu, J. *et al.* Any unique image biomarkers associated with COVID-19? *Eur. Radiol.* 1–7 (2020). doi:10.1007/s00330-020-06956-w

47. Amyar, A., Modzelewski, R. & Ruan, S. Multi-task Deep Learning Based CT Imaging Analysis For COVID-19: Classification and Segmentation. *medRxiv* 2020.04.16.20064709 (2020). doi:10.1101/2020.04.16.20064709

48. Ardakani, A. A., Kanafi, A. R., Acharya, U. R., Khadem, N. & Mohammadi, A. Application of deep learning technique to manage COVID-19 in routine clinical practice using CT images: Results of 10 convolutional neural networks. *Comput. Biol. Med.* **121**, 103795 (2020).

49. Mei, X. *et al.* Artificial intelligence–enabled rapid diagnosis of patients with COVID-19. *Nat. Med.* 1–5 (2020). doi:10.1038/s41591-020-0931-3

50. Han, Z. *et al.* Accurate Screening of COVID-19 Using Attention-Based Deep 3D Multiple Instance Learning. *IEEE Trans. Med. Imaging* **39**, 2584–2594 (2020).

51. Acar, E., ŞAhİN, E. & Yilmaz, İ. Improving effectiveness of different deep learning-based models for detecting COVID-19 from computed tomography (CT) images. *medRxiv* (2020). doi:10.1101/2020.06.12.20129643

52. Chen, X., Yao, L., Zhou, T., Dong, J. & Zhang, Y. Momentum Contrastive Learning for Few-Shot COVID-19 Diagnosis from Chest CT Images. *arXiv* (2020). Available at: http://arxiv.org/abs/2006.13276. (Accessed: 29th July 2020)

53. Jin, S. *et al.* AI-assisted CT imaging analysis for COVID-19 screening: Building and deploying a medical AI system in four weeks. *medRxiv* 1–22 (2020). doi:10.1101/2020.03.19.20039354

54. Wang, S. *et al.* A fully automatic deep learning system for COVID-19 diagnostic and prognostic analysis. *Eur. Respir. J.* **56**, (2020).

55. Shah, V. *et al.* Diagnosis of COVID-19 using CT scan images and deep learning techniques. *medRxiv* (2020). doi:10.1101/2020.07.11.20151332

56. Wang, J. *et al.* Prior-Attention Residual Learning for More Discriminative COVID-19 Screening in CT Images. *IEEE Trans. Med. Imaging* **39**, 2572–2583 (2020).

57. Wang, M. *et al.* Deep learning-based triage and analysis of lesion burden for COVID-19: a retrospective study with external validation. *Lancet Digit. Heal.* **2**, e506–e515 (2020).

58. Deng, J. *et al.* ImageNet: A large-scale hierarchical image database. in 248–255 (Institute of Electrical and Electronics Engineers (IEEE), 2010). doi:10.1109/cvpr.2009.5206848

59. Goodfellow, I. J. *et al. Generative Adversarial Nets*. (2014).

60. Shi, F. *et al.* Large-Scale Screening of COVID-19 from Community Acquired Pneumonia using Infection Size-Aware Classification. *arXiv* (2020). Available at: http://arxiv.org/abs/2003.09860. (Accessed: 29th July 2020)

61. Guiot, J. *et al.* Development and validation of an automated radiomic CT signature for detecting COVID-19. *medRxiv* (2020). doi:10.1101/2020.04.28.20082966

62. Chen, X. X. *et al.* A diagnostic model for coronavirus disease 2019 (COVID-19) based on radiological semantic and clinical features: a multi-center study. *Eur. Radiol.* 1–10 (2020). doi:10.1007/s00330-020-06829-2

63. Qin, L. *et al.* A predictive model and scoring system combining clinical and CT characteristics for the diagnosis of COVID-19. *Eur. Radiol.* (2020). doi:10.1007/s00330-020-07022-1

64. Xie, C. *et al.* Discrimination of pulmonary ground-glass opacity changes in COVID-19 and non-COVID-19 patients using CT radiomics analysis. *Eur. J. Radiol. Open* **7**, (2020).

65. Xu, M. *et al.* Accurately Differentiating COVID-19, Other Viral Infection, and Healthy Individuals Using Multimodal Features via Late Fusion Learning. *medRxiv* (2020). doi:10.1101/2020.08.18.20176776

66. Chassagnon, G. *et al.* AI-Driven CT-based quantification, staging and short-term outcome prediction of

COVID-19 pneumonia - v2. *medRxiv* (2020). doi:10.1101/2020.04.17.20069187

67. Ghosh, B. *et al.* A Quantitative Lung Computed Tomography Image Feature for Multi-Center Severity Assessment of COVID-19. *medRxiv* (2020). doi:10.1101/2020.07.13.20152231

68. Wei, W., Hu, X. wen, Cheng, Q., Zhao, Y. ming & Ge, Y. qiong. Identification of common and severe COVID-19: the value of CT texture analysis and correlation with clinical characteristics. *Eur. Radiol.* (2020). doi:10.1007/s00330-020-07012-3

69. Li, M. D. *et al.* Improvement and Multi-Population Generalizability of a Deep Learning-Based Chest Radiograph Severity Score for COVID-19. *medRxiv* (2020). doi:10.1101/2020.09.15.20195453

70. Li, M. D. *et al.* Automated Assessment and Tracking of COVID-19 Pulmonary Disease Severity on Chest Radiographs using Convolutional Siamese Neural Networks. *Radiol. Artif. Intell.* **2**, e200079 (2020).

71. Schalekamp, S. *et al.* Model-based Prediction of Critical Illness in Hospitalized Patients with COVID-19. *Radiology* 202723 (2020). doi:10.1148/radiol.2020202723

72. Wang, X. *et al.* Multi-Center Study of Temporal Changes and Prognostic Value of a CT Visual Severity Score in Hospitalized Patients with COVID-19. *Am. J. Roentgenol.* (2020). doi:10.2214/ajr.20.24044

73. Yip, S. S. F. *et al.* Performance and Robustness of Machine Learning-based Radiomic COVID-19 Severity Prediction. *medRxiv* (2020). doi:10.1101/2020.09.07.20189977

74. Goncharov, M. *et al.* CT-based COVID-19 Triage: Deep Multitask Learning Improves Joint Identification and Severity Quantification - v1. *arXiv* (2020).

75. Lassau, N. *et al.* AI-based multi-modal integration of clinical characteristics, lab tests and chest CTs improves COVID-19 outcome prediction of hospitalized patients - v2. *medRxiv* (2020). doi:10.1101/2020.05.14.20101972

76. Qi, X. *et al. Machine learning-based CT radiomics model for predicting hospital stay in patients with pneumonia associated with SARS-CoV-2 infection: A multicenter study*. *medRxiv* (Cold Spring Harbor Laboratory Press, 2020). doi:10.1101/2020.02.29.20029603

77. Zhu, X. *et al.* Joint Prediction and Time Estimation of COVID-19 Developing Severe Symptoms using Chest CT Scan. *arXiv* (2020). Available at: http://arxiv.org/abs/2005.03405. (Accessed: 29th July 2020)

78. Cohen, J. P. *et al.* Predicting covid-19 pneumonia severity on chest x-ray with deep learning. *arXiv* **12**, (2020).

79. Chao, H. *et al.* Integrative Analysis for COVID-19 Patient Outcome Prediction. *ArXiv* (2020).

80. Wu, Q. *et al.* Radiomics analysis of computed tomography helps predict poor prognostic outcome in COVID-19. *Theranostics* **10**, 7231–7244 (2020).

81. Zheng, Y. *et al.* Development and validation of a prognostic nomogram based on clinical and ct features for adverse outcome prediction in patients with covid-19. *Korean J. Radiol.* **21**, 1007–1017 (2020).

82. Chen, Y. *et al.* A quantitative and radiomics approach to monitoring ards in COVID-19 patients based on chest CT: A retrospective cohort study. *Int. J. Med. Sci.* **17**, 1773–1782 (2020).

83. Yue, H. *et al.* Machine learning-based CT radiomics method for predicting hospital stay in patients with pneumonia associated with SARS-CoV-2 infection: a multicenter study. *Ann. Transl. Med.* **8**, 859–859 (2020).

84. Bai, H. X. *et al.* AI Augmentation of Radiologist Performance in Distinguishing COVID-19 from Pneumonia of Other Etiology on Chest CT. *Radiology* 201491 (2020).

85. Amyar, A., Modzelewski, R., Li, H. & Ruan, S. Multi-task deep learning based CT imaging analysis for COVID-19 pneumonia: Classification and segmentation. *Comput. Biol. Med.* **126**, (2020).

86. Kermany, D. S. *et al.* Identifying Medical Diagnoses and Treatable Diseases by Image-Based Deep

Learning. *Cell* **172**, 1122-1131.e9 (2018).

87. Ezzat, D., Hassanien, A. E. & Ella, H. A. An optimized deep learning architecture for the diagnosis of COVID-19 disease based on gravitational search optimization. *Appl. Soft Comput. J.* (2020). doi:10.1016/j.asoc.2020.106742

88. COVID-19 | Radiology Reference Article | Radiopaedia.org. Available at: https://radiopaedia.org/articles/covid-19-4?lang=gb. (Accessed: 29th July 2020)

89. COVID-19 DATABASE | SIRM. Available at: https://www.sirm.org/en/category/articles/covid-19-database/. (Accessed: 29th July 2020)

90. CORONACASES.org by RAIOSS.COM. Available at: https://coronacases.org/. (Accessed: 30th July 2020)

91. Eurorad. Available at: https://www.eurorad.org/. (Accessed: 29th July 2020)

92. Chest X-Ray Images (Pneumonia) | Kaggle. Available at: https://www.kaggle.com/paultimothymooney/chest-xray-pneumonia. (Accessed: 29th July 2020)

93. Maguolo, G. & Nanni, L. A Critic Evaluation of Methods for COVID-19 Automatic Detection from X-Ray Images. *arXiv* (2020). Available at: http://arxiv.org/abs/2004.12823. (Accessed: 13th August 2020)

94. RSNA Pneumonia Detection Challenge | Kaggle. Available at: https://www.kaggle.com/c/rsna-pneumonia-detection-challenge. (Accessed: 29th July 2020)

95. Bachtiger, P., Peters, N. & Walsh, S. L. Machine learning for COVID-19—asking the right questions. *Lancet Digit. Heal.* **2**, e391–e392 (2020).

96. Banerjee, I. *et al.* Was there COVID-19 back in 2012? Challenge for AI in Diagnosis with Similar Indications. *arXiv* (2020). Available at: http://arxiv.org/abs/2006.13262. (Accessed: 4th August 2020)

97. Vickers, A. J. & Elkin, E. B. Decision curve analysis: A novel method for evaluating prediction models. *Med. Decis. Mak.* **26**, 565–574 (2006).

98. Lambin, P. *et al.* Radiomics: The bridge between medical imaging and personalized medicine. *Nat. Rev. Clin. Oncol.* **14**, 749–762 (2017).

99. Mongan, J., Moy, L. & Kahn, C. E. Checklist for Artificial Intelligence in Medical Imaging (CLAIM): A Guide for Authors and Reviewers. *Radiol. Artif. Intell.* **2**, e200029 (2020).

100. Collins, G. S., Reitsma, J. B., Altman, D. G. & Moons, K. G. M. Transparent reporting of a multivariable prediction model for individual prognosis or diagnosis (TRIPOD): The TRIPOD statement. *Ann. Intern. Med.* **162**, 55–63 (2015).

101. Wolff, R. F. *et al.* PROBAST: A Tool to Assess the Risk of Bias and Applicability of Prediction Model Studies. *Ann. Intern. Med.* **170**, 51 (2019).

102. Whiting, P. F. *et al.* QUADAS-2: A Revised Tool for the Quality Assessment of Diagnostic Accuracy Studies. *Ann. Intern. Med.* **155**, 529 (2011).

103. Zhang, C., Bengio, S., Hardt, M., Recht, B. & Vinyals, O. Understanding deep learning requires rethinking generalization. *5th Int. Conf. Learn. Represent. ICLR 2017 - Conf. Track Proc.* (2016).

104. Liberati, A. *et al.* The PRISMA statement for reporting systematic reviews and meta-analyses of studies that evaluate healthcare interventions: explanation and elaboration. *BMJ* **339**, (2009).

105. Covid-19 living Data. Available at: https://covid-nma.com/. (Accessed: 29th July 2020)

106. WHO | Pneumonia of unknown cause – China. Available at: https://www.who.int/csr/don/05-january-2020-pneumonia-of-unkown-cause-china/en/. (Accessed: 29th July 2020)

107. Veritas Health Innovation, M. A. Covidence systematic review software. Available at: https://www.covidence.org/. (Accessed: 12th August 2020)

108. Aviles-Rivero, A. I., Sellars, P., Schönlieb, C.-B. & Papadakis, N. GraphXCOVID: Explainable Deep Graph Diffusion Pseudo-Labelling for Identifying COVID-19 on Chest X-rays. (2020).

## Figures and Tables

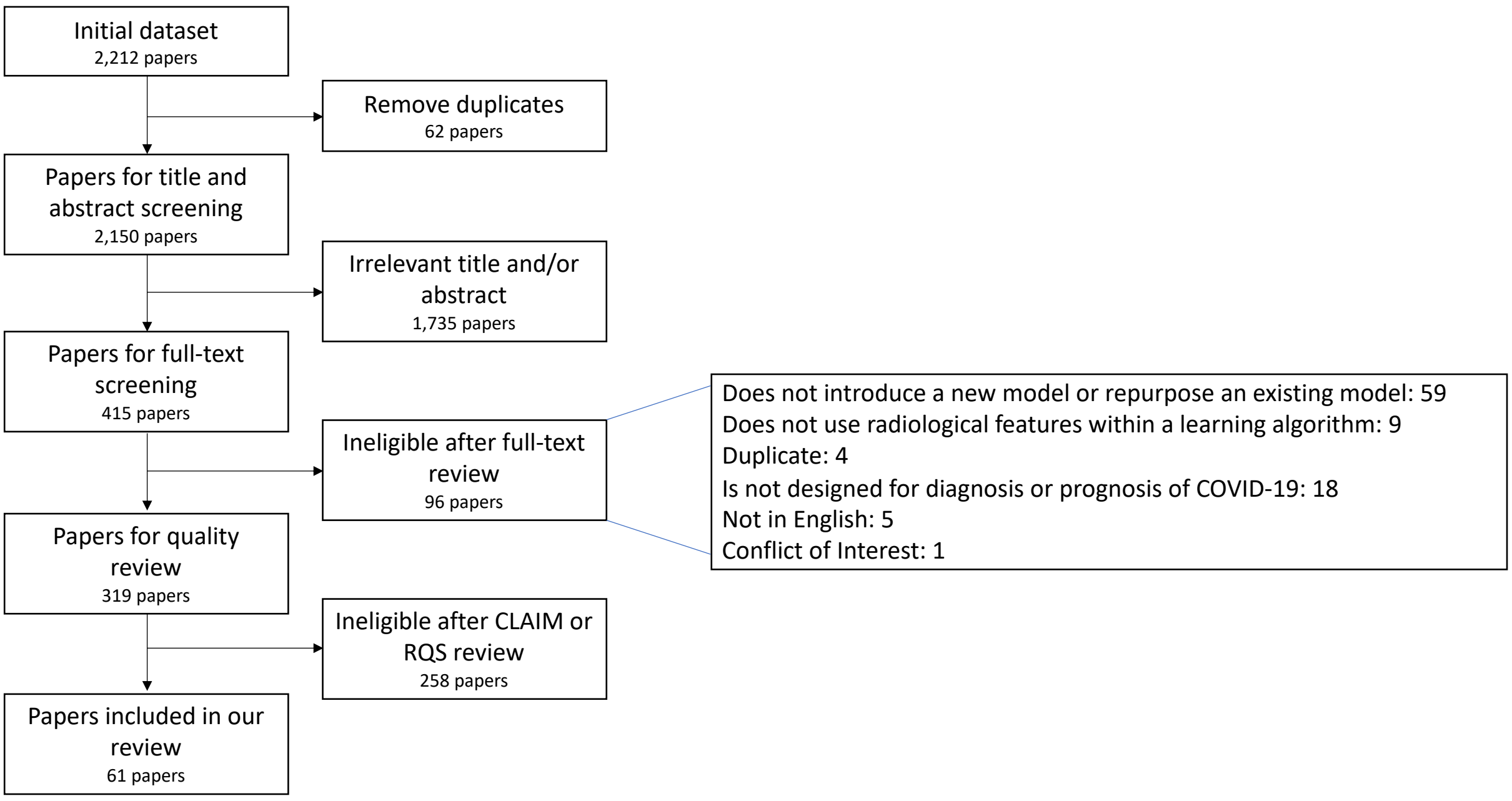

**Figure 1:** PRISMA flowchart for our systematic review, highlighting the inclusion and exclusion of papers at each stage. CLAIM: Checklist for Artificial Intelligence in Medical Imaging, RQS: Radiomic Quality Score.

| REFERENCE | DOMAIN | | | |
|---|---|---|---|---|
| | PARTICIPANTS | PREDICTORS | OUTCOMES | ANALYSIS |
| Ghoshal et al. [20] | HIGH | UNCLEAR (DL) | UNCLEAR | HIGH |
| Li et al. [37] | UNCLEAR | UNCLEAR (DL) | UNCLEAR | HIGH |
| Ezzat et al. [87] | HIGH | UNCLEAR (DL) | UNCLEAR | HIGH |
| Tartaglione et al. [19] | HIGH | UNCLEAR (DL) | UNCLEAR | HIGH |
| Luz et al. [33] | UNCLEAR | UNCLEAR (DL) | UNCLEAR | HIGH |
| Bassi et al. [34] | HIGH | UNCLEAR (DL) | UNCLEAR | HIGH |
| Kana et al. [35] | HIGH | UNCLEAR (DL) | UNCLEAR | HIGH |
| Heidari et al. [36] | HIGH | UNCLEAR (DL) | UNCLEAR | UNCLEAR |
| Farooq et al. [32] | HIGH | UNCLEAR (DL) | UNCLEAR | HIGH |
| Zhang et al. [30] | LOW | UNCLEAR (DL) | LOW | UNCLEAR |
| Zhang et al. [40] | HIGH | UNCLEAR (DL) | UNCLEAR | HIGH |
| Wang et al. [29] | HIGH | UNCLEAR (DL) | HIGH | HIGH |
| Bararia et al. [28] | HIGH | UNCLEAR (DL) | UNCLEAR | HIGH |
| Tsiknakis et al. [24] | HIGH | UNCLEAR (DL) | UNCLEAR | HIGH |
| Malhotra et al. [21] | HIGH | HIGH | UNCLEAR | HIGH |
| Sayyed et al. [39] | HIGH | LOW | UNCLEAR | HIGH |
| Rahaman et al. [22] | HIGH | UNCLEAR (DL) | UNCLEAR | HIGH |
| Amer et al. [23] | HIGH | UNCLEAR (DL) | UNCLEAR | HIGH |
| Elaziz et al. [25] | HIGH | LOW | UNCLEAR | HIGH |
| Tamal et al. [27] | HIGH | HIGH | UNCLEAR | HIGH |
| Gil et al. [26] | HIGH | LOW | UNCLEAR | UNCLEAR |
| Zokaeinikoo et al. [38] | HIGH | UNCLEAR (DL) | UNCLEAR | UNCLEAR |
| Amyar et al. [47] | HIGH | UNCLEAR (DL) | UNCLEAR | HIGH |
| Ardakani et al. [48] | HIGH | UNCLEAR (DL) | LOW | HIGH |
| Bai et al. [84] | HIGH | UNCLEAR (DL) | LOW | LOW |
| Jin et al. [53] | HIGH | UNCLEAR (DL) | LOW | UNCLEAR |
| Wang et al. [45] | HIGH | UNCLEAR (DL) | LOW | UNCLEAR |
| Ko et al. [44] | HIGH | UNCLEAR (DL) | HIGH | LOW |
| Acar et al. [51] | HIGH | UNCLEAR (DL) | HIGH | UNCLEAR |
| Pu et al. [46] | UNCLEAR | UNCLEAR (DL) | LOW | UNCLEAR |
| Chen et al. [52] | HIGH | UNCLEAR (DL) | UNCLEAR | UNCLEAR |
| Shah et al. [55] | HIGH | UNCLEAR (DL) | UNCLEAR | HIGH |
| Han et al. [50] | HIGH | UNCLEAR (DL) | UNCLEAR | HIGH |
| Wang et al. [56] | UNCLEAR | UNCLEAR (DL) | LOW | LOW |
| Wang et al. [57] | HIGH | UNCLEAR (DL) | HIGH | UNCLEAR |
| Goncharov et al. [74] | HIGH | UNCLEAR (DL) | HIGH | LOW |
| Xie et al. [64] | HIGH | LOW | LOW | HIGH |
| Xu et al. [65] | HIGH | LOW | LOW | UNCLEAR |
| Qin et al. [63] | LOW | HIGH | HIGH | HIGH |
| Georgescu et al. [43] | HIGH | UNCLEAR (DL) | HIGH | UNCLEAR |
| Guiot et al. [61] | HIGH | LOW | LOW | UNCLEAR |
| Shi et al. [60] | HIGH | LOW | LOW | LOW |
| Mei et al. [49] | HIGH | UNCLEAR (DL) | LOW | HIGH |
| Chen et al. [62] | HIGH | LOW | LOW | HIGH |
| Wang et al. [54] | UNCLEAR | UNCLEAR (DL) | LOW | LOW |
| Li et al. [69] | LOW | UNCLEAR (DL) | UNCLEAR | HIGH |
| Li et al. [70] | LOW | UNCLEAR (DL) | LOW | HIGH |
| Schalekamp et al. [71] | UNCLEAR | LOW | LOW | LOW |
| Cohen et al. [78] | HIGH | UNCLEAR (DL) | UNCLEAR | UNCLEAR |
| Yue et al. [83] | HIGH | LOW | HIGH | HIGH |
| Zhu et al. [77] | HIGH | LOW | LOW | LOW |
| Lassau et al. [75] | HIGH | UNCLEAR (DL) | HIGH | UNCLEAR |
| Chassagnon et al. [66] | UNCLEAR | LOW | LOW | UNCLEAR |
| Chao et al. [79] | LOW | LOW | LOW | HIGH |
| Wu et al. [80] | UNCLEAR | LOW | LOW | HIGH |
| Zheng et al. [81] | UNCLEAR | LOW | LOW | HIGH |
| Chen et al. [82] | HIGH | LOW | LOW | HIGH |
| Ghosh et al. [67] | HIGH | HIGH | HIGH | LOW |
| Wei et al. [68] | LOW | LOW | HIGH | HIGH |
| Wang et al.[72] | UNCLEAR | LOW | LOW | LOW |
| Yip et al. [73] | HIGH | LOW | UNCLEAR | UNCLEAR |

**Table 1:** PROBAST results for each domain considered for each paper included in our systematic review.

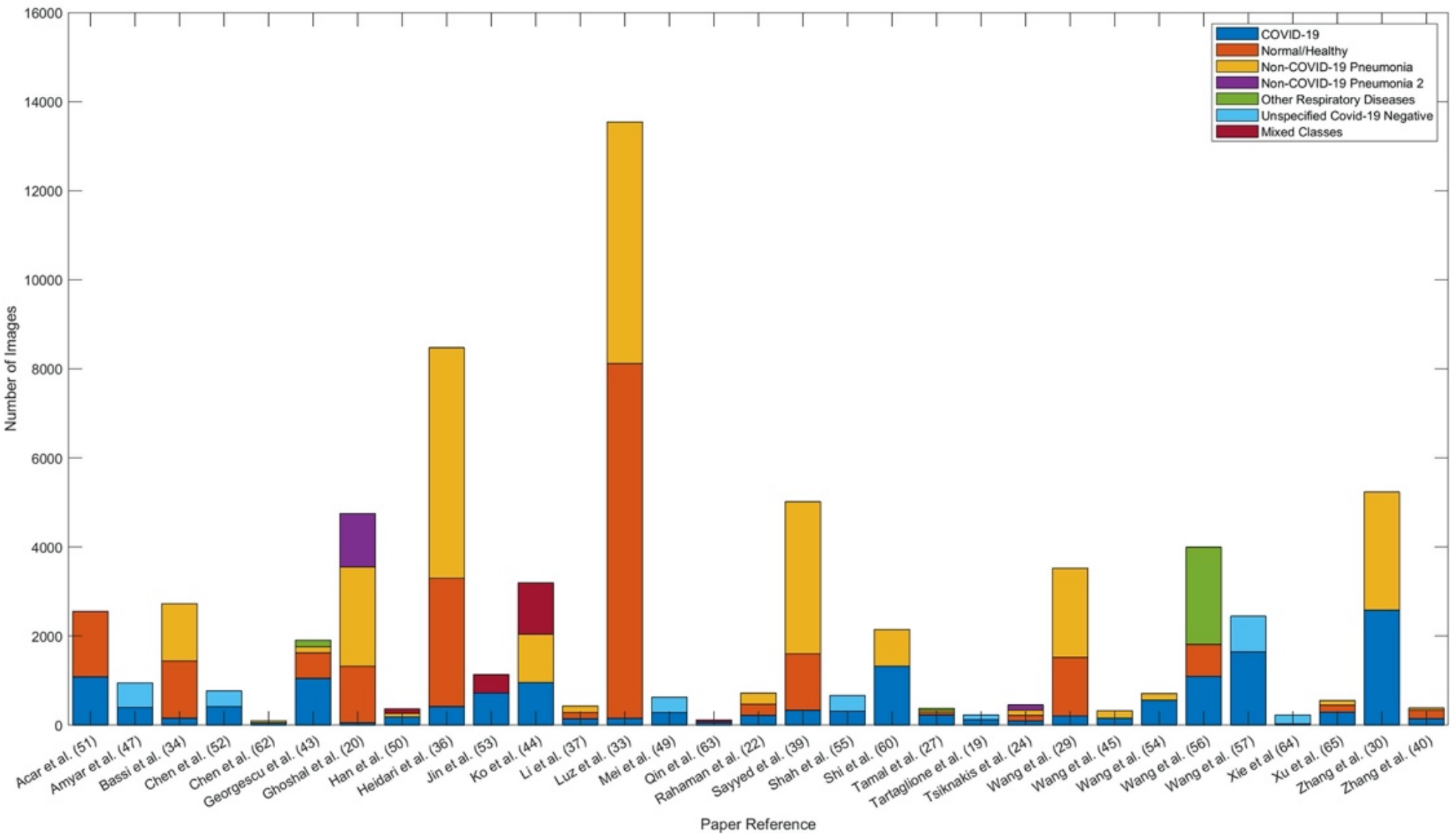

**Figure 2:** The number of images used in each paper for model training split by image class. Note: we exclude Bai et al.[84] from the figure as they use significantly more training data (118,401 images) than other papers and in Xu et al.[65] two COVID-19 classes are shown in the graph as one combined class.

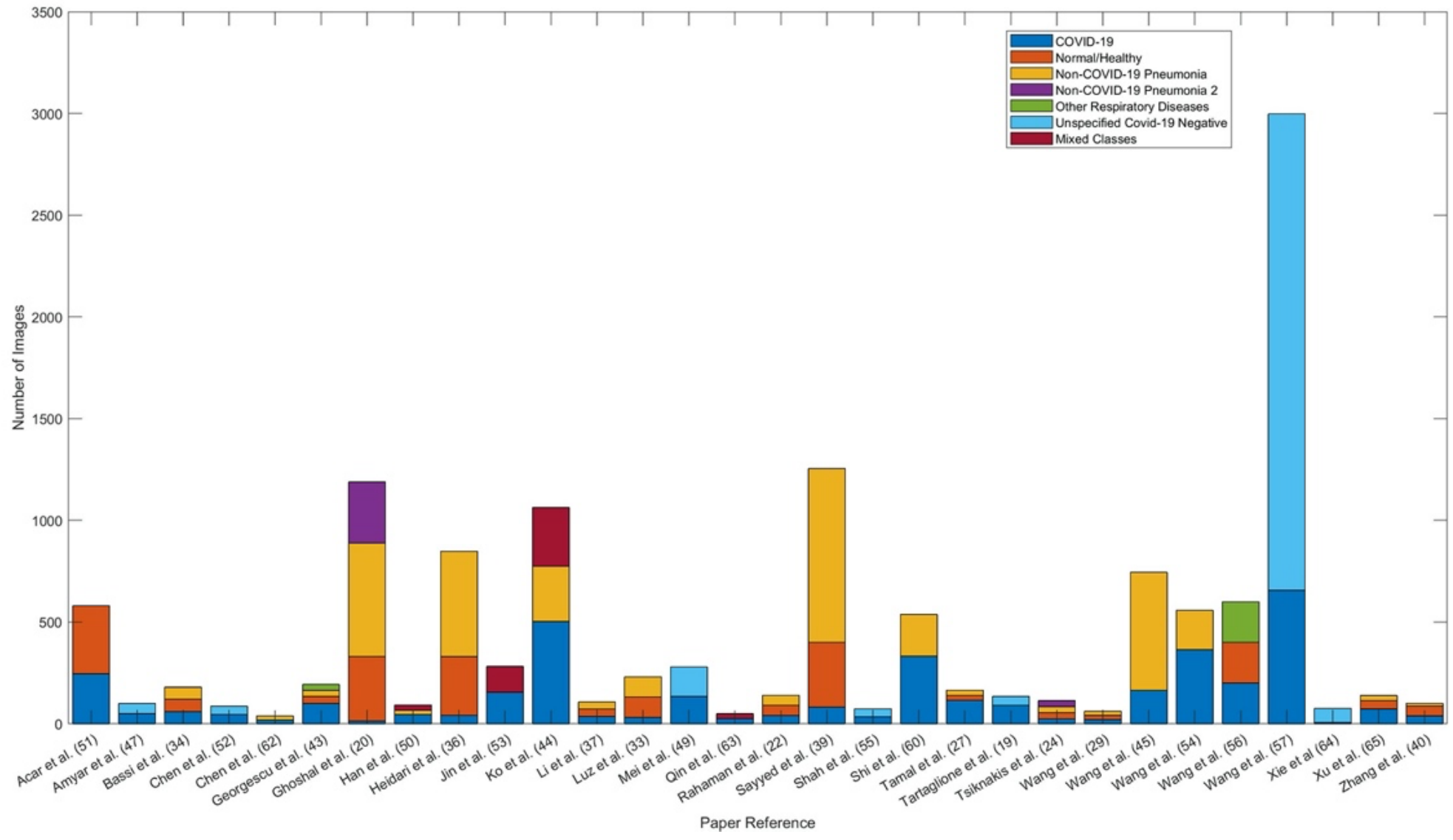

**Figure 3:** The number of images used for model testing split by image class. Note: we exclude Bai et al.[84] and Zhang et al.[30] from the figure as they use significantly more testing data (14,182 and 5869 images respectively) than other papers, and there are a large number of images (1,237) in the testing dataset in Wang et al.[57] which are unidentified in the paper (we include these in the unspecified COVID-19 negative).

## TABLE 2: DATA EXTRACTION

| REFERENCE | DIAGNOSIS / PROGNOSIS | DATA USED IN MODEL | PREDICTORS | SAMPLE SIZE DEVELOPMENT | SAMPLE SIZE TEST | TYPE OF VALIDATION | EVALUATION | PUBLIC CODE |
|---|---|---|---|---|---|---|---|---|
| | **Is this paper describing a COVID-19 diagnosis or prognosis model (or both?)** | **Does this use CXR or CT? (or both?)** | **What are the predictors? In purely deep-learning models, this is DL.** | **Total sample size used for development (i.e. training and validation and NOT test set), along with number of positive outcomes.** | **Total sample size used for testing of the algorithm, along with the number of positive outcomes.** | **k-fold CV, external validation in k centres, no validation, etc** | **Performance of the model, AUC, confidence interval, sensitivity, specificity, etc. 95% CI if available.** | **Is there code available ? (Is the trained model available ?)** |
| **Ghoshal et al. [20]** | Diagnosis | CXR | DL | 4,752 images, 54 COVID-19. | 1,189 images, 14 COVID-19. | Unclear validation procedure. | Unclear in the paper. | No |
| **Li et al. [37]** | Diagnosis | CXR | DL | 429 images, 143 COVID-19 | 108 images, 36 COVID-19. | Internal holdout validation. | Accuracy: 0·880, AUC: 0·970. | Yes (Yes) |
| **Ezzat et al. [87]** | Diagnosis | CXR | DL | Unclear in the paper. | Unclear in the paper. | Internal holdout validation. | Precision (w): 0·98, Recall (w): 0·98, F1 score (w): 0·98. ***(w): weighted average*** | No |
| **Tartaglione et al. [19]** | Diagnosis | CXR | DL | 231 images, 126 COVID-19. | 135 images, 90 COVID-19. | Internal holdout validation. | Unclear in the paper. | No |
| **Luz et al.** [33] | Diagnosis | CXR | DL | 13,569 images, 152 COVID-19. | 231 images, 31 COVID-19. | Internal holdout validation. | Accuracy = 0·94, Sensitivity: 0·97, PPV: 1·00. | Yes (Yes) |
| **Bassi et al.** [34] | Diagnosis | CXR | DL | 2724 images, 159 COVID-19. | 180 images, 60 COVID-19. | Internal holdout validation | Recall: 0·98, Precision: 1·00. | No |
| **Kana et al.** [35] | Diagnosis | CXR | DL | Unclear in the paper. | Unclear in the paper. | External validation. | Accuracy: 0·99, Recall: 1·00, Precision: 0·99, F1: 1·00. | No |
| **Heidari et al.** [36] | Diagnosis | CXR | DL | 8474 images, 415 COVID-19. | 848 images, 42 COVID-19. | Internal holdout validation. | Precision (w): 0·95, Recall (w): 0.94, F1-score (w): 0·94. (w): weighted average | No |
| **Farooq et al.** [32] | Diagnosis | CXR | DL | Unclear in the paper. | 637 images, 8 COVID-19. | Internal holdout validation. | Accuracy: 0.96, Sensitivity: 0.97, PPV: 0.99, F1: 0.98. | No |
| **Zhang et al.** [30] | Diagnosis | CXR | DL | 5236 images, 2582 COVID-19. | 5869 images, 3223 COVID-19. | Internal holdout validation. | AUC: 0.92, Sensitivity: 0.88, Specificity: 0.79. | Yes (No) |
| **Zhang et al.** [40] | Diagnosis | CXR | DL | 386 images, 150 COVID-19. | 101 images, 39 COVID-19. | Internal holdout validation. | Accuracy: 0.91. | No |
| **Wang et al.** [29] | Diagnosis | CXR | DL | 3522 images, 204 COVID-19. | 61 images, 20 COVID-19. | Internal holdout validation. | AUC: 1.00, Accuracy: 0.99. | No |
| **Bararia et al.** [28] | Diagnosis | CXR | DL | Unclear in the paper. | 1000 images, 341 COVID-19. | Internal holdout validation. | Accuracy: 0.81, Sensitivity: 0.81, Specificity: 0.90, Precision: 0.74, Recall: 0.77, F1: 0.75. | No |
| **Tsiknakis et al.** [24] | Diagnosis | CXR | DL | 458 (CV) images, 98 COVID-19. | 114 (CV) images, 24 COVID-19. | Five-fold internal cross-validation. | AUC: 1·00, Accuracy 1·00, Sensitivity: 0·99, Specificity: 1·00. | Yes (No) |
| **Malhotra et al.** [21] | Diagnosis | CXR | DL | 26,464 images, 1,740 COVID-19[#] | 6,299 images, 125 COVID-19[#] | Internal holdout validation. | Sensitivity: 0·87, Specificity: 0·97. | No |
| **Sayyed et al.** [39] | Diagnosis | CXR | DL | 5,018 (CV) images, 334 COVID-19. | 1,255 (CV) images, 83 COVID-19. | Five-fold internal cross-validation. | Accuracy: 0·99±0·05, | Yes (No) |
| **Rahaman et al.** [22] | Diagnosis | CXR | DL | 720 images, 220 COVID-19. | 140 images, 40 COVID-19. | Internal holdout validation. | Accuracy: 0·89, Precision: 0·90, Recall: 0·89, F1: 0·90. | No |
| **Amer et al.** [23] | Diagnosis | CXR | DL | Unclear in the paper. | Unclear in the paper. | Internal holdout validation. | AUC: 0·98, Accuracy: 0·94, Sensitivity: 0·92, Specificity: 0·97, PPV: 0·98. | No |
| **Elaziz et al.** [25] | Diagnosis | CXR | Hand-engineered radiomic features. | Unclear in the paper. | Unclear in the paper. | Internal holdout validation and external validation. | **Internal validation:** Accuracy: 0·96, Recall: 0·99, Precision: 0.96. **External validation:** Accuracy: 0·98, Recall: 0·99, Precision: 0·99. | No |
| **Tamal et al.** [27] | Diagnosis | CXR | Hand-engineered radiomic features. | 378 images, 226 COVID-19. | 165 images, 115 COVID-19. | Internal holdout validation. | Sensitivity: 1.00 Specificity: 0.85. | No* |
| **Gil et al.** [26] | Diagnosis | CXR | Hand-engineered radiomic features. | Unclear in the paper. | Unclear in the paper. | Internal holdout validation. | Accuracy: 0·96, Sensitivity: 0·98, Specificity: 0·93, Precision: 0·96. | Yes (Yes) |
| **Zokaeinikoo et al.** [38] | Diagnosis | CXR and CT | DL | Unclear in the paper. | Unclear in the paper. | Ten-fold internal cross-validation. | Accuracy: 0·99, Sensitivity: 0·99, Specificity: 1·00, PPV: 1·00 | No |
| **Amyar et al.** [47] | Diagnosis | CT | DL | 944 patients, 399 COVID-19. | 100 patients, 50 COVID-19. | Internal holdout validation. | Accuracy: 0·95, Sensitivity: 0·96, Specificity: 0·92, AUC: 0·97. | No |
| **Ardakani et al.[48]** | Diagnosis | CT | DL | Unclear as splits do not total correctly. | Unclear as splits do not total correctly. | Internal holdout validation | AUC: 0·99, Sensitivity, 1·00, Specificity, 0·99, Accuracy, 1·00; PPV, 0·99, NPV, 1·00. | No |
| **Bai et al. [84]** | Diagnosis | CT | DL | 118,401 images, 60,776 COVID-19. | 14,182 images, 5,030 COVID-19. | Internal holdout validation | AUC: 0·95, Accuracy: 0·96, Sensitivity: 0·95, Specificity: 0·96. | Yes (Yes) |
| **Jin et al. [53]** | Diagnosis | CT | DL | 1136 images, 723 COVID-19. | 282 images, 154 COVID-19. | Internal holdout validation. | Sensitivity: 0·97, Specificity: 0·92, AUC: 0·99. | No |
| **Wang et al. [45]** | Diagnosis | CT | DL | 320 images, 160 COVID-19. | Internal validation: 455 images, 95 COVID-19.<br><br>External validation: 290 images, 70 COVID-19. | Internal holdout validation and external validation. | **Internal validation:** AUC: 0·93 [0·90,0·96].<br><br>**External validation:** AUC: 0·81 [0·71,0·84]. | No |
| **Ko et al. [44]** | Diagnosis | CT | DL | 3,194 (CV) images, 955 COVID-19. | Internal cross-validation: 799 (CV) images, 239 COVID-19<br><br>External validation: 264 images, All COVID-19 | Five-fold internal cross-validation and external validation. | **Internal validation:** AUC: 1·00, Accuracy: 1.00, Sensitivity: 1.00, Specificity: 1·00.<br><br>**External validation:** Accuracy: 0·97. | No |
| **Acar et al.** [51] | Diagnosis | CT | DL | 2,552 images, 1,085 COVID-19. | 580 images, 246 COVID-19. | Internal holdout validation. | AUC: 1·00, Accuracy: 1·00, Error: 0·01, Precision: 1·00, Recall: 1·00, F1-Score: 1·00. | No |
| **Pu et al.** [46] | Diagnosis | CT | DL | Unclear in the paper. | Unclear in the paper. | Internal holdout validation | AUC: 0·70 [0·56,0·85], Sensitivity: 0·98, Specificity: 0·28. | No |
| **Chen et al.** [52] | Diagnosis | CT | DL | 770 (CV) images, 413 COVID-19. | Internal cross-validation: 86 (CV) images, 46 COVID-19 | Ten-fold internal cross-validation. | AUC: 0·94±0·01, Accuracy: 0·88±0·01, Precision: 0·90±0·01, Recall: 0·88±0·01. | No |
| **Shah et al.** [55] | Diagnosis | CT | DL | 664 images, 314 COVID-19. | 74 images, 35 COVID-19. | Internal holdout validation. | Accuracy: 0·95. | No |
| **Han et al.** [50] | Diagnosis | CT | DL | 368 (CV) images, 184 COVID-19. | 92 (CV) images, 46 COVID-19. | Five-fold internal cross-validation. | AUC: 0·99, Accuracy: 0·98. | No* |
| **Wang et al.** [56] | Diagnosis | CT | DL | 3997 images, 1095 COVID-19. | 600 images, 200 COVID-19. | Internal holdout validation. | AUC: 0.97, Accuracy: 0.93, Specificity: 0.96, Precision: 0.88, Recall: 0.88. | No |
| **Wang et al.** [57] | Diagnosis | CT | DL | 2447 images, 1647 COVID-19. | Internal validation: 639 images, 439 COVID-19.<br><br>External validation: 2120 images, 217 COVID-19. | Internal holdout and external validation. | **Internal validation:** AUC: 0.99, Sensitivity: 0.97, Specificity: 0.85.<br><br>**External validation:** AUC: 0.95, Sensitivity: 0.92, Specificity: 0.85. | No |
| **Goncharov et al.** [74] | Diagnosis and severity prognosis | CT | DL | Unclear in the paper. | Diagnosis: 101 images, 33 COVID-19.<br><br>Severity: 38 images of differing severity. | Internal holdout validation. | **Diagnosis model:** AUC: 0.95<br><br>**Severity model:** Correlation: 0.98 | No*** |
| **Xie et al.** [64] | Diagnosis | CT | Hand-engineered radiomic features | 225 images, 27 COVID-19. | 76 images, 6 COVID-19. | Internal holdout validation. | AUC: 0.91, Accuracy: 0.90, Sensitivity: 0.83, Specificity: 0.90. | No |
| **Xu et al.** [65] | Diagnosis | CT | DL and hand-engineered radiomic features | 551 images, 289 COVID-19. | 138 images, 73 COVID-19. | Internal holdout validation. | Accuracy: 0.98, F1: 0.99. | No*** |
| **Qin et al.** [63] | Diagnosis | CT | Hand-engineered radiomic features. | 118 patients, 62 COVID-19. | 50 patients, 26 COVID-19. | Internal holdout validation. | AUC: 0·85 [0·74, 0·96], Sensitivity: 0·89, Specificity: 0·92. | No |
| **Georgescu et al.** [43] | Diagnosis | CT | DL and hand-engineered radiomic features | 1,902 patients, 1,050 COVID-19. | 194 patients, 100 COVID-19. | Internal holdout validation. | AUC: 0·90, Sensitivity: 0·86, Specificity: 0·81. | No |
| **Guiot et al.** [61] | Diagnosis | CT | Hand-engineered radiomic features. | Unclear in the paper. | Unclear in the paper. | Internal holdout validation. | AUC = 0·94 [0·88,1·00], Accuracy: 0·90 [0·84, 0·94], Sensitivity: 0·79, Specificity: 0·91. | No |
| **Shi et al.** [60] | Diagnosis | CT | Hand-engineered radiomic features | 2,148 (CV) images, 1,326 COVID-19. | Internal cross-validation: 537 (CV) images, 332 COVID-19 | Five-fold internal cross-validation. | AUC: 0·94, Accuracy 0·88, Sensitivity: 0·91, Specificity: 0·83. | No |
| **Mei et al.** [49] | Diagnosis | CT | DL and CNN extracted features and clinical data | 626 images, 285 COVID-19. | 279 images, 134 COVID-19. | Internal holdout validation. | AUC = 0·92 [0·89,0·95], Sensitivity: 0·843 [0·77, 0·90], Specificity: 0·83 [0·76, 0·89]. | Yes (Yes) |
| **Chen et al.** [62] | Diagnosis | CT | Clinical features, qualitative imaging features and hand-engineered radiomic imaging features | 98 patients, 51 COVID-19. | 38 images, 19 COVID-19. | Internal holdout validation. | AUC: 0·94 [0·87,1·00], Accuracy: 0·76, Sensitivity: 0·74, Specificity: 0·79. | No |
| **Wang et al.** [54] | Diagnosis and prognosis for length of hospital stay. | CT | Diagnosis model: DL<br><br>Prognosis model: 64 CNN features and clinical factors. | 709 images, 560 COVID-19. | Validation 1: 226 images, 102 COVID-19.<br><br>Validation 2: 161 images, 92 COVID-19.<br><br>Validation 3: 53 images, All with length of hospital stay.<br><br>Validation 4: 117 images, All with length of hospital stay. | External validation. | **Validation 1 (Diagnosis):** AUC: 0·87<br>**Validation 2 (Diagnosis):** AUC: 0·88<br>**Validation 3 (Prognosis):** KM separation p = 0·01<br>**Validation 4 (Prognosis):** KM separation p = 0·01 | Yes (Yes) |
| **Li et al.** [69] | Prognosis for severity. | CXR | DL | 354 images of differing severities. | Internal validation: 108 images.<br><br>External validation: 111 images. | Internal holdout validation and external validation. | **Internal validation:** Correlation: 0.88.<br><br>**External validation:** Correlation: 0.90. | Yes (No) |
| **Li et al.** [70] | Prognosis for severity. | CXR | DL | 314 images of differing severities. | Internal validation: 154 images.<br><br>External validation: 113 images. | Internal holdout validation and external validation. | **Internal validation:** Correlation: 0.86.<br><br>**External validation:** Correlation: 0.86. | Yes (No) |
| **Schalekamp et al.** [71] | Prognosis for severity. | CXR | Hand-engineered radiomic features and clinical factors. | Unclear in the paper. | Unclear in the paper. | Internal holdout validation. | AUC: 0.77. | No |
| **Cohen et al.** [78] | Prognosis of lung opacity and extent of lung involvement with GGOs for COVID-19 patients. | CXR | Features from a trained CNN extracted at various layers. | 47 patients of varying severity. | 47 patients of varying severity. | Internal holdout validation. | Opacity correlation: 0·80<br>Extent correlation: 0·78 | Yes (Yes) |
| **Yue et al.** [83] | Prognosing short and long-term (> 10 day) hospital stay in for COVID-19 patients | CT | Hand-engineered radiomic features. | 26 patients,<br>16 long-term. | Internal validation<br>5 patients,<br>3 long-term.<br><br>Temporal-split internal validation:<br>6 patients,<br>All long-term. | Internal holdout and temporal-split validation. | AUC: 0·97 [0·83,1·00],<br>Sensitivity: 1·00,<br>Specificity: 0·89,<br>NPV: 1·00,<br>PPV: 0·80. | Yes** |
| **Zhu et al.** [77] | The prognosis for whether patients will convert to a severe stage of COVID-19 and regression to predict the time to that conversion. | CT | Hand-engineered radiomic features | Unclear in the paper | Unclear in the paper | Five-fold internal cross-validation run 20 times, average reported. | AUC: 0·86±0·02<br>Accuracy: 0·86±0·02<br>Sensitivity: 0·77±0·03<br>Specificity: 0·88±0·015 | No |
| **Lassau et al.** [75] | The prognostic model used for predicting the risk of death, need for ventilation or requirement for over 15L/min oxygen. | CT | CNN extracted features and clinical data | 646 patients,<br>All COVID-19.<br>243 with severe outcomes. | Internal validation:<br>150 images,<br>All COVID-19,<br>48 with severe outcome.<br><br>External validation:<br>135 patients,<br>All COVID-19,<br>unclear number of severe patients. | Internal holdout validation and external validation. | **Internal validation:**<br>AUC: 0·76<br><br>**External validation:**<br>AUC: 0·75 | No*** |
| **Chassagnon et al.** [66] | Short-term prognosis intubation and death within 4 days<br><br>Long-term prognosis: death within one month after CT | CT | Hand-engineered radiomic features and clinical data. | 536 COVID-19 patients,<br>108 severe short-term outcomes, unclear for long-term. | 157 COVID-19 patients,<br>31 severe short-term outcomes, unclear for long-term. | External validation. | **<u>Short-term prognosis:</u>**<br>Precision (w): 0·94,<br>Sensitivity (w): 0·94,<br>Specificity (w): 0·81,<br>Balanced Accuracy: 0·88.<br><br>**<u>Long-term prognosis:</u>**<br>Precision (w): 0·77,<br>Sensitivity (w): 0·94,<br>Specificity (w): 0·82,<br>Balanced Accuracy: 0·71.<br><br>***<u>(w): weighted</u>*** | No* |
| **Chao et al.** [79] | Prognosing for ICU admission | CT | Hand-engineered radiomic features and clinical data. | 236 (CV) images, 125 admitted to ICU. | 59 (CV) images, 31 admitted to ICU. | Five-fold internal cross-validation. | Unclear in the paper. | No |
| **Wu et al.** [80] | Prognosing for death, ventilation and ICU admission in early and late stage COVID-19. | CT | Hand-engineered radiomic features. | 351 images, 25 severe outcomes. | 141 images, 26 severe outcomes. | External validation. | **<u>Early stage COVID-19:</u>**<br>AUC: 0·86,<br>Sensitivity: 0·80,<br>Specificity: 0·86.<br><br>**<u>Late stage COVID-19:</u>**<br>AUC: 0·98,<br>Sensitivity: 1·00.<br>Specificity: 0·94. | No |
| **Zheng et al.** [81] | Prognosing for admission to an ICU, use of mechanical ventilation, or death. | CT | Hand-engineered radiomic features and clinical data. | 166 images, 35 severe outcomes. | 72 images, 10 severe outcomes. | External validation. | C-index: 0·89. | No |
| **Chen et al.** [82] | Prognosis for acute respiratory distress syndrome. | CT | Hand-engineered radiomic features and clinical data. | 247 images, 36 severe cases. | 105 images, 15 severe cases. | Internal holdout validation. | Accuracy: 0·88, Sensitivity: 0·55, Specificity: 0·95. | No |
| **Ghosh et al.** [67] | Prognosing COVID-19 severity. | CT | Hand-engineered radiomic features. | 36 images, unclear number of severe cases. | 24 images, unclear number of severe cases. | Internal holdout validation. | Accuracy: 0·88. | No |
| **Wei et al.** [68] | Prognosing COVID-19 severity. | CT | Hand-engineered radiomic features. | Unclear in the paper | Unclear in the paper | One-hundred fold leave-group-out cross validation. | AUC: 0·93 Accuracy: 0·91, Sensitivity: 0·81, Specificity: 0·95. | No |
| **Wang et al.** [72] | Prognosis for survival. | CT | Hand-engineered radiomic features. | 161 patients, 15 non-survivors. | 135 patients, unclear number of non-survivors. | External validation. | C-index: [0.92, 0.95], Accuracy: [0.85, 0.87], Sensitivity: [0.71,0.76], Specificity: [0.91,0.92]. | No |
| **Yip et al.** [73] | Prognosing COVID-19 severity. | CT | Hand-engineered radiomic features. | 657 images of various severities. | 441 images of various severities. | Internal holdout validation. | AUC: 0.85. | No |

*** the authors state that the algorithm will be made publicly available.**
**** the authors state that “imaging or algorithm data used in this study are available upon request.”**
***** the paper states that code “is available on a public GitHub repository” but no link is provided and the authors could not locate it.**
**# number of samples after augmentation, the original number of COVID-19 images is unclear.**

**Table 2:** Summary of the data extracted for each paper included in our systematic review

## Appendix

### 1. Search strategy

**Initial extraction.** For the arXiv papers, we initially extract papers for the relevant date ranges that include “ncov” (as a complete word), “coronavirus”, “covid”, “sars-cov-2” or “sars-cov2" in their title or abstract.For the “Living Evidence on COVID-19” [105], we download all papers in the appropriate date range.

**Refined search.** We then filter the identified papers using the following criteria: title or abstract contain one of: “ai”,“deep”,”learning”, “machine”, “neural”, “intelligence”, “prognos”, “diagnos”, “classification”, “ segmentation” and also contain one of “ct”, “cxr”, “x-ray”, “xray”, “imaging”, “image”, “radiograph”. Only “ai”, “ct” and “cxr” are required to be complete words.

This differs slightly from the original PROSPERO description; the search was widened to identify some additional papers. The full history of the searching and filtering source code can be explored at https://gitlab.developers.cam.ac.uk/maths/cia/covid-19-systematic-review

### 2. Supplementary author list

| Name | Affiliation |
|---|---|
| Alessandro Ruggiero | Royal Papworth Hospital, Cambridge, UK; Qureight Ltd, Cambridge, UK |
| Anna Korhonen | Language Technology Laboratory, University of Cambridge, Cambridge, UK |
| Emily Jefferson | Population Health and Genomics, School of Medicine, University of Dundee, Dundee, UK |
| Emmanuel Ako | Chelsea and Westminster NHS Trust and Royal Brompton NHS Hospital, London, UK |
| Georg Langs | Department of Biomedical Imaging and Image-guided Therapy, Computational Imaging Research Lab Medical University of Vienna, Vienna, Austria |
| Ghassem Gozaliasl | Department of Physics, University of Helsinki, Helsinki, Finland |
| Guang Yang | National Heart & Lung Institute, Imperial College London, London, UK |
| Helmut Prosch | Department of Biomedical Imaging and Image-Guided Therapy, Medical University of Vienna, Austria; Boehringer Ingelheim |
| Jacobus Preller | Addenbrooke’s Hospital, Cambridge University Hospitals NHS Trust, Cambridge, UK |
| Jan Stanczuk | Department of Mathematics and Theoretical Physics, Cambridge University, Cambridge, UK |
| Jing Tang | Research Program in System Oncology, Faculty of Medicine, University of Helsinki, Helsinki, Finland |
| Johannes Hofmanninger | Department of Biomedical Imaging and Image-guided Therapy, Computational Imaging Research Lab Medical University of Vienna, Vienna, Austria |
| Judith Babar | Addenbrooke’s Hospital, Cambridge University Hospitals NHS Trust, Cambridge, UK |
| Lorena Escudero Sánchez | Department of Radiology, University of Cambridge, UK; Cancer Research UK Cambridge Institute, Cambridge, UK |
| Muhunthan Thillai | Interstitial Lung Disease Unit, Royal Papworth Hospital, Cambridge, UK; Department of Medicine, University of Cambridge, Cambridge, UK |
| Paula Martin Gonzalez | Cancer Research UK Cambridge Centre, University of Cambridge, Cambridge, UK |
| Philip Teare | Biopharmaceuticals R&D, AstraZeneca, Cambridge, UK |
| Xiaoxiang Zhu | Signal Processing in Earth Observation, Technical University of Munich, Munich, Germany |
| Mishal Patel | Biopharmaceuticals R&D, AstraZeneca, Cambridge, UK |
| Conor Cafolla | Department of Chemistry, University of Cambridge, Cambridge, UK |
| Hojjat Azadbakht | AINOSTICS Ltd, Manchester, UK |
| Joseph Jacob | Centre for Medical Image Computing, University College London, London, UK |
| Josh Lowe | SparkBeyond UK Ltd, London, UK |
| Kang Zhang | Center for Biomedicine and Innovations at Faculty of Medicine, Macau University of Science and Technology, Macau, China |
| Kyle Bradley | SparkBeyond UK Ltd, London, UK |
| Marcel Wassin | contextflow GmbH, Vienna, Austria |
| Markus Holzer | contextflow GmbH, Vienna, Austria |
| Kangyu Ji | Cavendish Laboratory, University of Cambridge, Cambridge, UK |
| Maria Delgado Ortet | Department of Radiology, Cambridge University, Cambridge, UK |
| Tao Ai | Tongji Hospital, Tongji Medical College, Huazhong University of Science and Technology, Wuhan, China |
| Nicholas Walton | Institute of Astronomy, University of Cambridge, Cambridge, UK |
| Pietro Lio | Department of Computer Science and Technology, University of Cambridge, Cambridge, UK |
| Samuel Stranks | Department of Chemical Engineering and Biotechnology, University of Cambridge, Cambridge, UK |
| Tolou Shadbahr | Research Program in Systems Oncology, Faculty of Medicine, University of Helsinki, Helsinki, Finland |
| Weizhe Lin | Department of Engineering, University of Cambridge, Cambridge, UK |
| Yunfei Zha | Department of Radiology, Renmin Hospital of Wuhan University, Wuhan, China |
| Zhangming Niu | Aladdin Healthcare Technologies Ltd, London, UK |